\documentclass[a4paper,fleqn]{cas-dc}

\usepackage[numbers]{natbib}
\usepackage{algorithm}
\usepackage{algpseudocode}

\def\tsc#1{\csdef{#1}{\textsc{\lowercase{#1}}\xspace}}
\tsc{WGM}
\tsc{QE}
\tsc{EP}
\tsc{PMS}
\tsc{BEC}
\tsc{DE}

\begin{document}
\let\WriteBookmarks\relax
\def\floatpagepagefraction{1}
\def\textpagefraction{.001}
\shorttitle{BRSR-OpGAN}
\shortauthors{M.U. Zahid et~al.}

\title [mode = title]{BRSR-OpGAN: Blind Radar Signal Restoration using Operational Generative Adversarial Network}                      



\author[1]{Muhammad Uzair Zahid}
\cormark[1]

\credit{Conceptualization of this study, Methodology, Software}

\affiliation[1]{organization={Department of Computing Sciences, Tampere University},
                city={Tampere},
                postcode={33100}, 
                country={Finland}}

\author[2]{Serkan Kiranyaz}
\cormark[1]
\affiliation[2]{organization={Department of Electrical Engineering,
College of Engineering, Qatar University, Qatar},
                city={Doha},
                postcode={2713}, 
                country={Qatar}}

\author[3]{Alper Yildirim}
\affiliation[3]{organization={Department of Electrical and Electronics Engineering, School Of Engineering and Natural Sciences, Ankara Medipol University},
                city={Ankara},
                postcode={06050}, 
                country={Turkey}}


\author[1]{Moncef Gabbouj}

\cortext[cor1]{Corresponding authors. \\
E-mail addresses: muhammaduzair.zahid@tuni.fi (M.U. Zahid), mkiranyaz@qu.edu.qa (S. Kiranyaz).}


\begin{abstract}
Many studies on radar signal restoration in the literature focus on isolated restoration problems, such as denoising over a certain type of noise, while ignoring other types of artifacts. Additionally, these approaches usually assume a noisy environment with a limited set of fixed signal-to-noise ratio (SNR) levels. However, real-world radar signals are often corrupted by a blend of artifacts, including but not limited to unwanted echo, sensor noise, intentional jamming, and interference, each of which can vary in type, severity, and duration. This study introduces Blind Radar Signal Restoration using an Operational Generative Adversarial Network (BRSR-OpGAN), which uses a dual domain loss in the temporal and spectral domains. This approach is designed to improve the quality of radar signals, regardless of the diversity and intensity of the corruption. The BRSR-OpGAN utilizes 1D Operational GANs, which use a generative neuron model specifically optimized for blind restoration of corrupted radar signals. This approach leverages GANs' flexibility to adapt dynamically to a wide range of artifact characteristics. The proposed approach has been extensively evaluated using a well-established baseline and a newly curated extended dataset called the Blind Radar Signal Restoration (BRSR) dataset. This dataset was designed to simulate real-world conditions and includes a variety of artifacts, each varying in severity. The evaluation shows an average SNR improvement over 15.1 dB and 14.3 dB for the baseline and BRSR datasets, respectively. Finally, the proposed approach can be applied in real-time, even on resource-constrained platforms. This pilot study demonstrates the effectiveness of blind radar restoration in time-domain for real-world radar signals, achieving exceptional performance across various SNR values and artifact types. The BRSR-OpGAN method exhibits robust and computationally efficient restoration of real-world radar signals, significantly outperforming existing methods.
\end{abstract}




\maketitle

\section{Introduction}
Radar Signal restoration is an integral process that enhances the detection and classification of radar signals. This process involves removing noise and interference from the received signal, which can be caused by various sources, such as atmospheric conditions, system imperfections, and external sources.  The target signal can be detected and identified more accurately by effectively restoring the received signal, allowing for more precise measurements and classifications. This is particularly important in applications such as weather forecasting, electronic warfare (EW), and vehicle safety, where the reliability and accuracy of various radar systems are critical. In EW, the detection and classification of threat radar signals is of vital importance \cite{de2018introduction}. EW receivers aim to detect radars from the longest possible distance. Low probability of intercept (LPI) radars are challenging to detect by EW receivers due to their low output power and ability to spread their energy across the frequency spectrum \cite{pace2009detecting}. The SNR improvement, i.e., the signal processing gain, is vital to facilitate the detection of LPI radars from longer distances. Therefore, radar signal restoration is an integral part of radar signal processing and is an active area of research in remote sensing and radar signal processing.

\begin{figure*}[t!]
    \centering
    \includegraphics[width=1\textwidth,keepaspectratio]{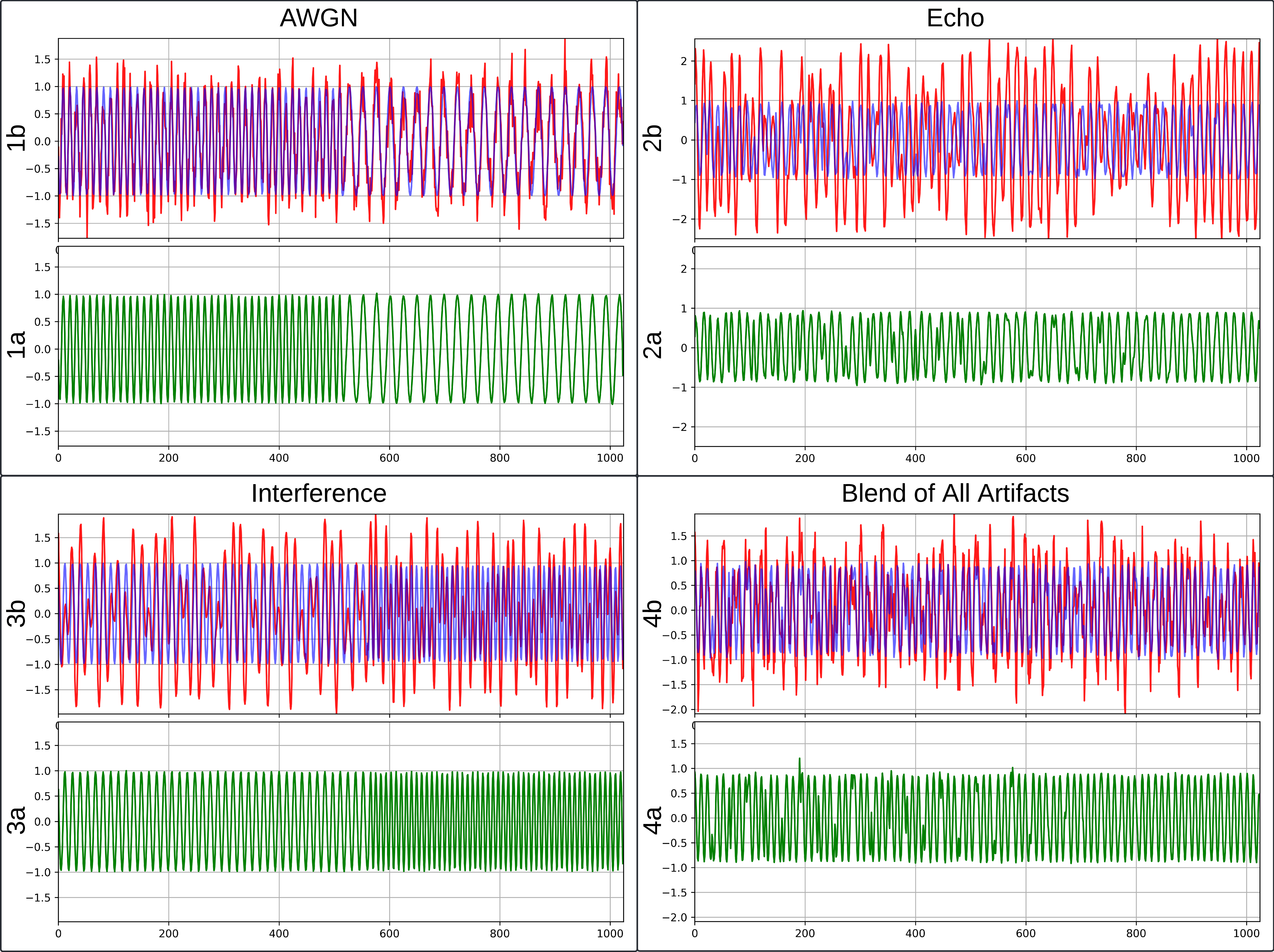}
    \caption{Sample signals from the extended dataset demonstrating the impact of various artifacts and the effectiveness of the proposed restoration method. In each example, the clean signal (blue) and the corrupted signal (red) are overlaid in the top plot (a), while the restored signal (green) is shown in the bottom plot (b). Common artifacts include Additive White Gaussian Noise (AWGN), Echo, Interference, and a blend of all with randomized weights.}
    \label{fig:figure_1}
\end{figure*}

However, as the number of radar radiation sources increases, the electromagnetic environment becomes increasingly complex. This complexity often results in disturbances in the radar signals captured by the receiver, leading to a relatively low SNR that presents significant challenges for recognizing radar signals. Existing methods in radar signal restoration often target specific distortion types with fixed SNR levels, focusing on isolated tasks such as denoising or echo reduction. Despite their effectiveness in controlled environments, these traditional techniques are insufficient for dealing with the complexity of real-world situations. 

Real-world scenarios pose a confluence of signal corruption challenges, including unwanted echo, interference, and sensor noise such as Additive White Gaussian Noise (AWGN), varying in type, severity, and duration as illustrated in Fig.~\ref{fig:figure_1}. Given the complex and variable nature of these distortions, restoring radar signals requires a comprehensive and blind approach capable of addressing the broad spectrum of signal corruption scenarios typically encountered in real-world radar environments. Essentially, such an approach must go beyond the limitations of conventional methods and offer robust solutions that can adapt to the diverse and dynamic nature of practical radar signals.

Time-frequency analysis is a crucial technique for extracting intra-pulse features of radar emitters, transforming signals into time-frequency images (TFIs). Notable methods in time-frequency analysis encompass the short-time Fourier transform (STFT) \cite{zhou2019ensemble}, Wigner-Ville distribution (WVD) \cite{kishore2017automatic}, and Choi-Williams distribution (CWD) \cite{seddighi2020radar}, among others. These methods have gained popularity for their ability to provide a detailed representation of signals across both time and frequency domains. Upon generating TFIs, recent advancements in radar waveform recognition have increasingly incorporated deep learning (DL) to analyze these images, extracting features and identifying waveforms with enhanced precision \cite{jan2020artificial,zhou2018automatic,zhang2017convolutional}. Multilayer denoising is proposed in \cite{jiang2024multilayer} using a sequence of variational mode decomposition (VMD), local mean decomposition (LMD), and wavelet thresholding (WT) methods for the three-stage decomposition of the noisy radar signal. The processed signal was then transformed into a time-frequency image using the Choi-Williams Distribution (CWD). Subsequently, a neural network with dilated convolution was trained on time-frequency images, achieving a recognition rate of 75.3\% under low SNR conditions. Similarly,  \cite{lee2019mutual} presented a wavelet denoising approach for suppressing mutual interference in automotive FMCW radar systems, demonstrating significant enhancements in target detection accuracy. 

Recent advancements in deep learning techniques have paved the way for innovative solutions to radar signal denoising. Fuchs et al. \cite{fuchs2020automotive} explored automotive radar interference mitigation using a convolutional autoencoder, highlighting the effectiveness of deep learning models in denoising and improving signal-to-noise-plus-interference ratios. Previous works, such as \cite{wang2017automatic,kong2018automatic,zhang2019automatic}, use time-frequency analysis methods to transform raw radar signals into 2D images, following which deep networks are utilized, which significantly increase computational complexity.
DNCNet was introduced in \cite{du2022dncnet}, a deep neural network designed for efficient denoising and classification of radar signals. They utilized two subnetworks: a denoising subnetwork using a U-Net architecture with a noise level estimation module and a classification subnetwork. The authors proposed a two-phase training procedure where the denoising subnetwork is initially trained, followed by a phase to strengthen the mapping between denoising results and perceptual representation. Although direct evaluations of the restored signals were not provided, the study focused on the classification performance of the signals across different SNR levels, demonstrating significant improvements in classification performance over the restored signals.

The advent of Generative Adversarial Networks (GANs) \cite{goodfellow2014generative} has opened new avenues in the field of signal processing, offering innovative ways to tackle complex restoration tasks \cite{mvuh2024multichannel}. GANs, through their adversarial training mechanism, have shown remarkable success in generating high-quality outputs in various domains, including signal, image, and speech processing. This success prompts the exploration of their applicability in the domain of radar signal restoration, particularly in a one-dimensional (1D) context, which is inherently suited for processing spatio-temporal signals such as radar signals. Operational GANs (OpGANs) represent a recent evolution of the GAN framework that enables the model to learn more robust and accurate mappings from degraded to clean signals by incorporating 1D Self-ONN \cite{ince2022blind,kiranyaz2022blind} layers into the GAN architecture, allowing for handling a more comprehensive set of signal degradations, addressing real-world artifacts' diversity and unpredictability.

Building on these insights, in this study, we aim to tackle this problem by using a blind restoration approach without making any prior assumptions about the type and severity of each artifact corrupting the signal. We propose a novel BRSR-OpGAN model that incorporates compact Self-Organized Operational Neural Networks (Self-ONNs) \cite{kiranyaz2021self} for both the generator and discriminator. Previous studies have demonstrated that Self-ONNs outperform CNNs in various regression and classification tasks due to their efficient computational processing and generative neuron model \cite{zahid2021robust, gabbouj2022robust, zahid2022global}. Despite having only 11 layers in the generator network and 7 layers in the discriminator network, this study demonstrates that relatively shallow OpGAN networks can achieve an unprecedented blind restoration performance over real-world radar signals. We also introduce a novel loss function for the generator, which incorporates spectrogram-based loss terms in addition to the time domain loss, improving the generator's output waveform in both time and frequency domains. Additionally, we use 1D-SelfONN layer-based residual blocks to further enhance the restoration performance.

The lack of benchmark datasets and rigorous evaluation protocols for radar signal restoration remains a significant challenge in the field. Datasets currently used in the literature are accompanied by disclaimers on their respective websites, indicating that they may not be suitable for comprehensive evaluation \cite{DeepSig}. Moreover, existing studies predominantly rely on classification performance metrics to assess the effectiveness of signal restoration techniques. This focus on classification accuracy, both pre- and post-restoration, complicates direct comparisons of signal quality improvement across different approaches. Consequently, there is an urgent need to develop standardized datasets and evaluation frameworks prioritizing signal quality metrics to enable more meaningful and consistent benchmarking of radar signal restoration methods. To generate a large-scale real-world radar signal dataset,  the BRSR benchmark dataset with a random blend of artifacts is compiled with 85,800 radar modulation signals from the 12-class dataset referenced in \cite{jiang2024multilayer}. These signals were then corrupted by randomly selected artifacts each with randomly varying weights. The applied radar artifacts include AWGN noise, a set of 100 signals for interference, and an unwanted echo model with randomized attenuation and delay. The random combination and artifact severity resulted in radar corruption at different SNR levels, ranging from $-14$ to $10$ dB. 

This study addresses a critical gap in radar signal restoration datasets and presents the true potential of 1D OpGANs for blind signal restoration. Through rigorous testing and validation, our research underscores the importance of innovative approaches in overcoming the challenges of signal degradation, marking a pioneering advancement in the field of radar signal processing. The novel and significant contributions of this study can be summarized as follows:

\begin{enumerate}
  \item This pilot study approaches the radar signal restoration blindly, i.e., no assumptions were made regarding the types and severities of artifacts on the received radar signal.
  \item In this study, we present the first application of 1D OpGANs to the restoration of a radar signal in real-time by employing the raw signals directly.
  \item In order to enhance the restoration performance of the generator, a loss function was developed, leveraging both time and frequency domain information.
  \item Unlike previous studies, this research focuses on restoring various types of distortions, particularly those that are severely corrupted, while the SNR randomly varies between -14 dB and 10 dB.
  \item To simulate real-world radar signals with any combination of artifacts and severities, an extended BRSR dataset has been curated with paired clean and corrupted signals and is publicly shared along with the source codes.
  \item The proposed approach has been evaluated on benchmark datasets, achieving an exceptional level of average SNR restoration performance, surpassing 15 dB for the baseline and over 14 dB for the BRSR dataset. Furthermore, the proposed BRSR-OpGAN encapsulates a compact generator model with an elegant computational complexity, allowing real-time processing on a single CPU computer.
\end{enumerate}

The rest of the paper is organized as follows: Section \ref{sec:methodology} briefly outlines 1D Self-ONNs, the proposed approach with BRSR-OpGAN model architecture, and the experimental setup. Section \ref{sec:dataset} presents the baseline data set used and the creation of a benchmark BRSR dataset. Section \ref{sec:results} presents the results over a wide range of restoration experiments and evaluates the performances achieved. Finally, Section \ref{sec:conclusion} concludes the paper and suggests avenues for future research.

\section{METHODOLOGY}
\label{sec:methodology}
In this section, we begin by summarizing the fundamentals of Self-Organized Operational Neural Networks (Self-ONNs) and their primary attributes. Subsequently, we outline our novel method for utilizing 1D OpGANs for radar signal restoration.

\subsection{1D Self-Organized Operational Neural Networks}
In this section, we provide a brief description of the main network characteristics of 1D Self-ONNs. Unlike conventional CNNs with linear convolution operation and fixed (static) nodal operators in ONNs\cite{kiranyaz2020operational,kiranyaz2021exploiting,kiranyaz2017generalized,kiranyaz2017progressive}, Self-ONNs feature generative neurons with the ability to adopt any arbitrary nodal function, denoted as $\psi$, (including standard types such as linear and harmonic functions) that are optimized for each kernel element and connection. This operational flexibility and diversity of Self-ONNs enable them to form optimized nodal operator functions during training, thus maximizing the learning performance.

In the Self-ONN, each kernel element performs a nonlinear transformation, $\psi$, whose function can be expressed using the Taylor series near the origin (a=0), 
\begin{equation}
\psi(x) = \sum_{n=0}^{\infty} \frac{\psi^{(n)}(0)}{n!} x^n
\end{equation}

The  $Q^{th}$ order truncated approximation, formally known as the \textit{MacLaurin} polynomial, takes the form of the following finite summation:
\begin{equation}
\psi(x)^{(Q)} = \sum_{n=0}^{Q} \frac{\psi^{(n)}(0)}{n!} x^n
\label{eq:2}
\end{equation}

Any nonlinear function $\psi(x)$ near zero can be approximated by the above formulation. When the activation function bounds the neuron’s input feature maps in the vicinity of 0 (e.g., tanh), the formulation of \eqref{eq:2} can be exploited to form a composite nodal operator where the power coefficients, $\frac{\psi^{(n)}(0)}{n!}$, can be the parameters of the network learned during training. 

It was shown in \cite{gabbouj2022robust} that the nodal operator of the kth generative neuron in the $l^{th}$ layer can take the following general form:
\begin{multline}
\widetilde{\psi_k^l} \left( w_{ik}^{l(Q)}(r), y_i^{l-1}(m + r) \right) \\
= \sum_{q=1}^{Q} w_{ik}^{l(Q)}(r, q) \left( y_i^{l-1}(m + r) \right)^q
\end{multline}

Let $x_{ik}^l \in \mathbb{R}^M$ denote the contribution of the $i^{th}$ neuron at the $(l-1)^{th}$ layer to the input map of the $l^{th}$ layer, which can be articulated as:
\begin{equation}
\widetilde{x_{lk}^l}(m) = \sum_{r=0}^{K-1} \sum_{q=1}^{Q} w_{ik}^{l(Q)}(r, q) \left( y_i^{l-1}(m + r) \right)^q
\label{eq:4}
\end{equation}

In this formulation, $y_i^{l-1} \in \mathbb{R}^M$ is the output map of the $i^{th}$ neuron at the $(l-1)^{th}$ layer, and $w_{ik}^{l(Q)}$ represents a learnable kernel matrix of dimensions $K \times Q$, i.e., $w_{ik}^{l(Q)} \in \mathbb{R}^{K \times Q}$, formed as,  
\begin{equation}
w_{ik}^{l(Q)}(r) = \left[ w_{ik}^{l(Q)}(r, 1), w_{ik}^{l(Q)}(r, 2), \ldots, w_{ik}^{l(Q)}(Q) \right]
\end{equation}

The commutativity of summation operations in \eqref{eq:4} allows for an alternative representation:
\begin{equation}
\widetilde{x_{lk}^l}(m) = \sum_{q=1}^{Q} \sum_{r=0}^{K-1} w_{ik}^{l(Q)}(r, q-1) \left( y_i^{l-1}(m + r) \right)^q
\end{equation}
which can be simplified as:
\begin{equation}
\widetilde{x_{lk}^l} = \sum_{q=1}^{Q} \text{Conv1D} \left( w_{ik}^{l(Q)}, \left( y_i^{l-1} \right)^q \right)
\end{equation}

Hence, the formulation can be accomplished by applying $Q$ 1D convolution operations. Finally, the output of this neuron can be formulated as follows:
\begin{equation}
x_k^l = b_k^l + \sum_{i=0}^{N_{(l-1)}} x_{ik}^l
\end{equation}

\begin{figure*}[t!]
    \centering
    \includegraphics[width=1\textwidth,keepaspectratio]{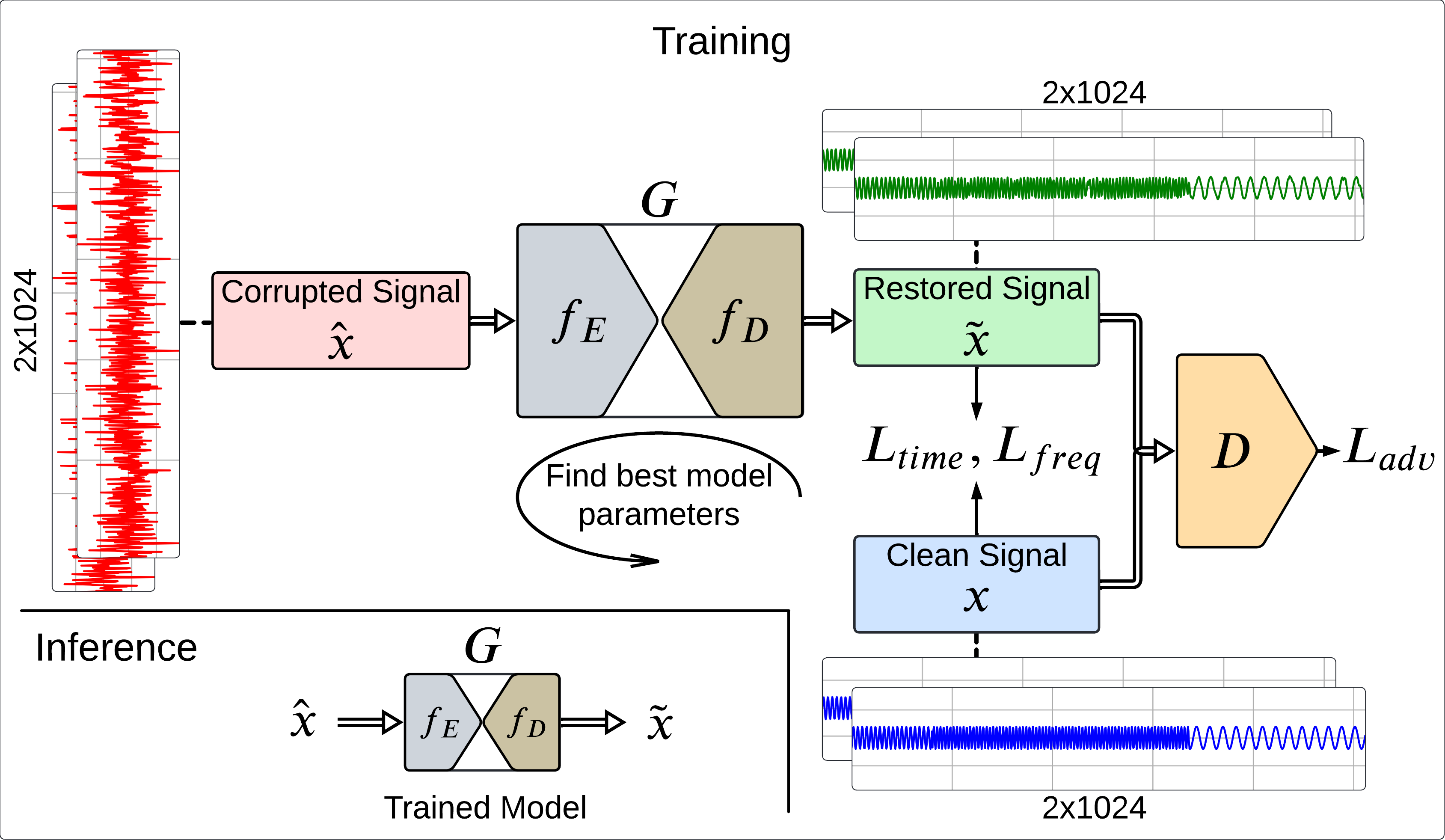}
    \caption{The proposed blind radar restoration approach using BRSR-OpGAN. After training, the encoder and decoder networks are employed together as a composite filter \( G = f_D \cdot f_E \) for denoising unseen radar signals.}
    \label{fig:figure_2}
\end{figure*}

where $b_k^l$ is the bias associated with the neuron. The $0^{th}$ order term (q=0), representing the DC bias, is omitted as the neuron's learnable bias parameter can compensate for each individual bias term. With the $Q=1$ setting, a generative neuron reduces back to a convolutional neuron.

The efficient formulations of the forward propagation and Backpropagation (BP) training in a raw-vectorized form can be found in \cite{malik2020fastonn}. 

\subsection{Proposed Approach}
The proposed approach for blind radar signal restoration using the BRSR-OpGAN model leverages a generative model architecture using paired corrupted (received) and clean (transmitted) radar signals. The primary objective of this study is to develop a supervised neural network adept at removing any blend of artifacts from the received radar signals. Utilizing a paired signal training dataset \(\{ (\hat{x}, x)_i \}_{i=1}^N\), where \(\hat{x}\) denotes the distorted version and \(x\) the clean radar signal for each frame, we aim to fine-tune the network's parameters to effectively map distorted inputs \(\hat{x}\) to their clean counterparts \(x\). Once these parameters are optimized, the generator network of the BRSR-OpGAN is capable of restoring the radar signals. As detailed earlier, our approach integrates 1D Self-ONN layers into the GAN's generator and discriminator, specifically designed to process complex valued radar signal waveforms. The overall framework of our proposed radar signal restoration method is depicted in Fig.~\ref{fig:figure_2}.

The proposed BRSR-OpGAN model executes time-domain, segment-based restoration on complex-valued radar signals, each with a length of \(1024\) samples and \(2\) channels where in-phase and quadrature (I/Q) samples are placed separately in each channel. To ensure unbiased training across different artifact types and severities, each segment is deliberately corrupted with a random blend of artifacts, including AWGN, echo, and interference. Each artifact is randomized at varying severity levels, controlled by randomly assigned weights as detailed in Section~\ref{sec:dataset}. The primary goal is for the trained model to consistently transform a corrupted segment (\(\hat{x}\)) into a clean segment (\(x\)), independent of its content, type, and severity of each artifact present as well as the modulation type. 

As illustrated in Fig.~\ref{fig:figure_2}, a raw radar signal segment from each batch is randomly selected as the input pair for the BRSR-OpGan. The normalization process applied to each segment is volume-independent, ensuring the model's performance is not biased by signal amplitude variations. The normalization is performed independently for each channel (I/Q) as follows:

\[
X_N^s(i) = \frac{X^s(i)}{X_{\text{max}}^s}
\]

where \( X^s(i) \) represents the original signal value at the \(i\)-th point in segment \( s \), \( X_{\text{max}}^s \) is the absolute maximum value in segment \( s \), and \( X_N^s(i) \) is the normalized signal value. This formula ensures that each segment is scaled relative to its absolute maximum value, thus standardizing the input for more effective processing by the BRSR-OpGAN.

\subsubsection{Model Architecture}
The proposed approach involves two main networks. The first is the generator network \( G \), consisting of an encoder \( f_E \) and a decoder \( f_D \), depicted in grey and brown, respectively, in Fig.~\ref{fig:figure_2}. These sub-networks work together to restore the signals during inference. The socond network, highlighted in orange, is the discriminator network \( D \), which plays a crucial role in the overall architecture.

The generator and discriminator networks are both 1-D Self-ONNs. Furthermore, the encoder network is specifically developed to process a 2-channel Radar signal, accommodating the real and imaginary parts of complex-valued radar signals. This design allows us to take advantage of the spatial information from each channel, leading to better noise reduction and signal clarity while preserving the radar components' morphology.

\begin{figure*}[t!]
    \centering
    \includegraphics[width=1\textwidth,keepaspectratio]{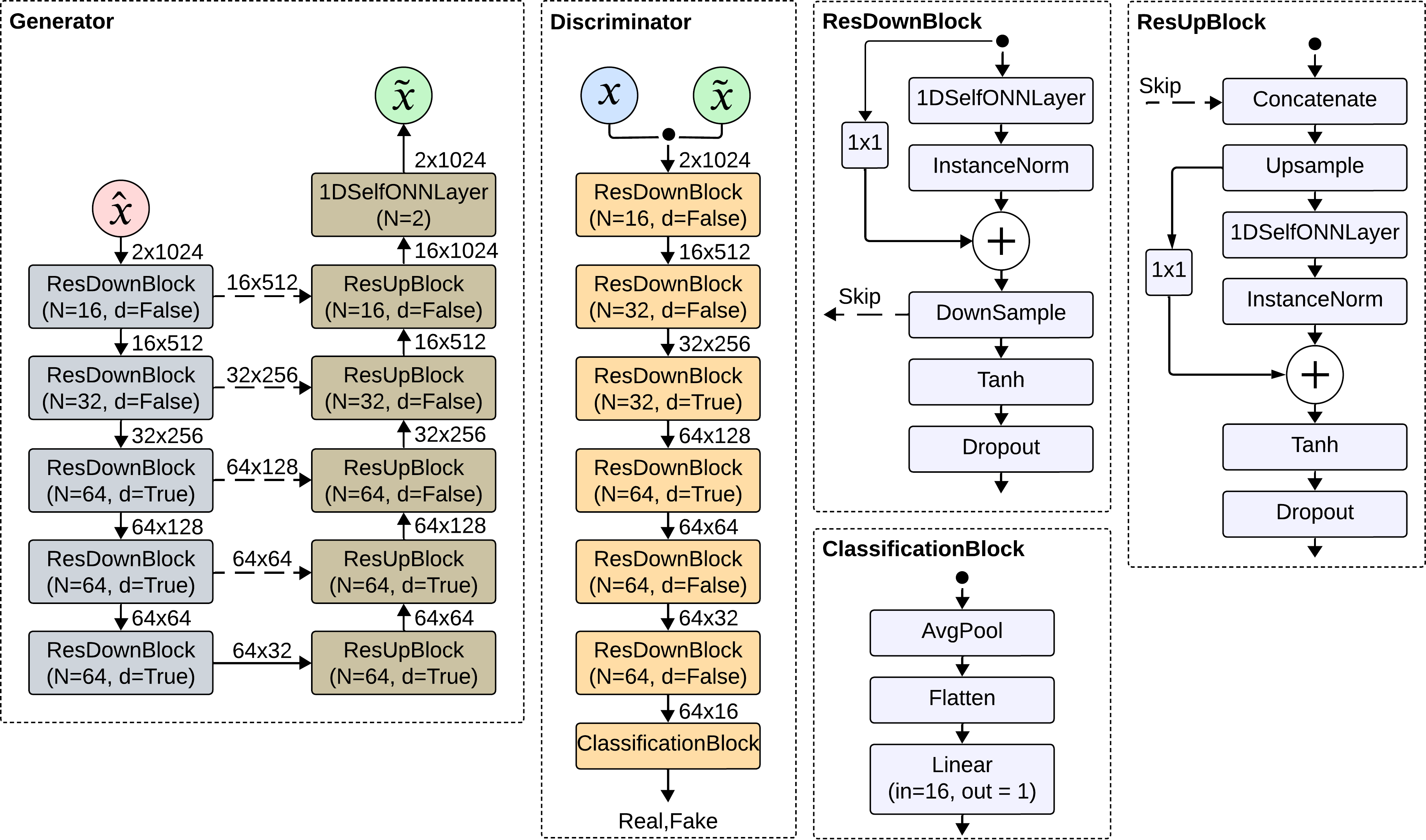}
    \caption{The generator and discriminator architectures of the BRSR-OpGAN. The ResDownBlock, ResUpBlock, and ClassificationBlock components are also detailed.}
    \label{fig:figure_3}
\end{figure*}

The encoder network \( f_E \) comprises five Residual downsampling blocks (ResDownBlock) blocks, each executing a sequence of four operations. These operations include a residual block utilizing 1-D Self-ONNs, instance normalization, downsampling, and a hyperbolic tangent (tanh) activation function. The encoder processes the corrupted radar signal \(\hat{x}\) and compresses it into a lower-dimensional vector \(\hat{z}\), which captures the essential features of \(x\) and is known as the latent vector of \(x\). The compression performed by \(f_E\) retains only the useful features of the signal, effectively filtering out noise.

Conversely, the decoder network \( f_D \) takes the latent representation \(\hat{z}\) and scales it back to a signal \(\tilde{x}\) that matches the dimensions of \(x\), making the direct comparison possible. It is composed of six blocks, the first five being residual downsampling blocks (ResDownBlock) that replicate the sequence of operations found in the encoder blocks, except that the downsampling is substituted with an upsampling operation. Finally, the decoder's final block performs a convolution operation without activation or normalization. Both the generated signal \(\tilde{x}\) and the clean signal \(x\) are fed into the discriminator. This process assists the generator in enhancing the signal quality by training it to produce signals that are indistinguishable from the clean signals, ultimately aiming to fake the discriminator.

The discriminator network \( D \) consists of six ResDownBlock, each primarily performing a downsampling operation using residual blocks similar to those in the decoder's downsampling blocks, except for the last one. The final block is the classification block, responsible for classifying the generated signal \(\tilde{x}\) against the clean signal \(x\). This classification block comprises an adaptive average pooling layer followed by a dense layer, which categorizes the output as either real or fake. This configuration is crucial for the adversarial learning component of the proposed approach, where \( D \) actively challenges \( G \) to produce increasingly realistic outputs.

Fig.~\ref{fig:figure_3} provides a comprehensive overview of the architectures for the encoder, decoder, and discriminator networks. Each ResDownBlock block in the encoder and its corresponding ResUpBlock block in the decoder are interconnected by dashed arrows, which act as skip connections. These skip connections are crucial for mitigating the problem of vanishing gradients often encountered in GAN architectures such as the one proposed. The figure also displays the input dimensionality of features at each layer, as well as the number of filters/neurons for each layer, denoted by \(N\). Additionally, some specific layers incorporate dropout (with a rate of 0.25), indicated as \(d=True\). The kernel sizes in the 1D Self-ONN layers are uniformly set to \(1 \times 3\).

\subsubsection{Objective Function}
Our approach utilizes an objective function similar to conditional Generative Adversarial Networks (cGANs) for image-to-image translation \cite{isola2017image}, tailored explicitly for training with paired noisy and clean radar signals. This structure facilitates a cGAN setup where the generator (denoiser) is trained to generate outputs that depend on the noisy inputs, intending to match the clean reference signals. This conditioning ensures that the generated outputs are statistically plausible and specific reconstructions of the input noise profiles.

A key innovation in our model is the inclusion of both time and frequency domain losses in the reconstruction loss component, which provides a more holistic fidelity assessment than standard approaches that only focus on the time domain. This dual-domain strategy allows the model to correct discrepancies effectively across both temporal and spectral characteristics, which is crucial for maintaining the integrity of radar signals.

This formulation ensures that the generator is trained to
produce outputs that are both realistic (as judged by the
discriminator) and accurate (as measured by the reconstruction
error). The objective function for training our model is formulated as follows:
\begin{multline}
 L_{tot}(G, D, \hat{x}, x) = \alpha L_{adv}(G, D, \hat{x}, x) \\
 + \beta \cdot L_{time}(G, D, \hat{x}, x) + \gamma \cdot L_{freq}(G, D, \hat{x}, x) 
\label{eq:total_loss}
\end{multline}
where \( L_{tot} \) is the total loss for the generator, \(L_{adv}\) represents the adversarial loss, which measures how indistinguishable the denoised outputs are from real, clean signals in terms of adversarial criteria, \(L_{time}\) represents the time-domain reconstruction loss, specifically an L1 loss that measures the mean absolute difference between the clean and the restored signals, \(L_{freq}\) is the frequency-domain reconstruction loss, involving the L1 loss between the normalized spectrograms of the clean and restored signals obtained via Short-Time Fourier Transform (STFT).

The coefficients \( \alpha \), \( \beta \), and \( \gamma \) are tuning parameters that balance the relative importance of each loss component within the overall objective function. These parameters are adjustable and should be optimized based on empirical performance during validation phases to ensure the best fidelity and perceptual quality combination in the restored radar signals.

In GAN training, the generator (G) and discriminator (D) networks are trained simultaneously in an adversarial manner. Specifically, the weights are updated based on competing objectives for the generator and discriminator: the generator seeks to minimize the loss function \( L_A(G, D, \hat{x}, x) \), while the discriminator seeks to maximize the corresponding loss function \( L_A(G, D, \hat{x}, x) \). In this setup, the generator is tasked with transforming the corrupted (received) signal \(\hat{x}\) into the target (transmitted) version \(\tilde{x}\). \(x\). The adversarial training ensures that the generator produces highly accurate and indistinguishable signal from the target signal over time.

\subsubsection{Training Procedure of the Generator Networks}
In the BRSR-OpGAN model, we focus on a structured training approach that leverages adversarial training alongside domain-specific reconstruction metrics to optimize performance. Here, we define the adversarial loss and dual-domain reconstruction losses utilized in our model.

\textbf{Adversarial Loss:} This component encourages the generator to produce outputs that the discriminator \(D\) will classify as real (close to clean signals). This loss is designed to decrease the discriminator's ability to distinguish generated signals from real signals:
\begin{equation}
L^{\star}_{adv}(D,\tilde{x}) = (D(\tilde{x}) - 1)^2 
\label{eq:adv_gen_loss}
\end{equation}

Here, the generator \(G\) tries to minimize this loss by generating signals \((\tilde{x})\) that are classified by \(D\) as real, thus aiming for \(D(G(\hat{x}))\) to be close to 1.

\textbf{Reconstruction Loss:} To ensure that the generated signals can fake the discriminator and accurate reconstructions of the target signals can be accomplished, we integrate both time-domain and frequency-domain metrics based on the L1 norm. This dual-domain approach ensures content accuracy across different representations of the signal. 

\textbf{Time Domain Reconstruction Loss:} This loss measures the accuracy of the generated signal \( \tilde{x}\) in replicating the real, clean signal \(x\) in the time domain:
\begin{equation}
     L^{\star}_{\text{time}}(\tilde{x},x) = \| \tilde{x} - x \|_1
\label{eq:time_loss}
\end{equation}

where \(\| \cdot \|_1\) denotes the L1 norm, which is the sum of the absolute differences between the generated signal and the true clean signal.

\textbf{Frequency Domain Performance Measurement:} In addition to the time domain loss, this component evaluates the accuracy of the generated signal in the frequency domain:
\begin{equation}
    L^{\star}_{\text{freq}}(\tilde{x}, x) = \left\| S(\tilde{x}) - S(x) \right\|_1
\label{eq:freq_loss}
\end{equation}

Where \(S(\cdot)\) represents the spectrogram obtained by the squared magnitude of STFT, which converts the time domain signals into their frequency components. The L1 norm is used to measure the absolute differences between the spectrograms of the generated and the real (target) signals.

The N-point discrete Short-Time Fourier Transform (STFT) is computed as follows:
\begin{equation}
    \text{STFT}[X, w, n] = X(n, w) = \sum_m X[m] W[n-m] e^{-jwm}
\label{eq:stft}
\end{equation}

where the N-point discrete STFT, which is complex-valued, is obtained by sampling at each discrete radial frequency, \(w = \frac{2\pi k}{N}\).

The spectrogram \( S(X) \) is then calculated as:
\begin{equation}
    S(X) = \left| \text{STFT}(X) \right|^2
\label{eq:spectrogram}
\end{equation}

\begin{algorithm}[t!]
\caption{Training BRSR-OpGAN Model}\label{algo1}
\label{alg:training}
\small
\begin{algorithmic}[1] 
\State \textbf{Input:} Training samples $\{\hat{x}_i, x_i\}_{i=1}^N$, balance coefficients $\alpha$, $\beta$, maximum iterations \text{maxIter}, learning rates $\eta_G$, $\eta_D$.
\State \textbf{Output:} Trained generator and discriminator weights: $\Theta_G, \Theta_D$.

\State Initialize trainable parameters, $\Theta_G, \Theta_D$;
\State Normalize the training samples $\hat{x}_i$ and $x_i$;

\While{$t < \text{maxIter}$}
    \State Sample a batch $(\hat{x}, x)$ from the training dataset;
    
    \Comment{Generator Training}
    \State Generate denoised output $\tilde{x} = G(\hat{x})$;
    \State Compute: \(L_1 = L^{\star}_{adv}(D,\tilde{x})\) \eqref{eq:adv_gen_loss};
    \State Compute: \(L_2 = L^{\star}_{\text{time}}(\tilde{x},x)\) \eqref{eq:time_loss}; 
    \State Compute: $L_3 = L^{\star}_{\text{freq}}(\tilde{x},x)$ \eqref{eq:freq_loss}
    \State Compute total generator loss:
    \[
    L_G = L_1 + \alpha L_2 + \beta L_3
    \]
    \State Update generator weights $\Theta_G$ using Adam optimizer to minimize $L_G$.

    \Comment{Discriminator Training}
    \State Recompute $\tilde{x} = G(\hat{x})$ using updated $\Theta_G$.
    \State Compute: $L_D = L_{adv}(D, \tilde{x}, x)$ \eqref{eq:disc_loss}

    \State Update discriminator weights $\Theta_D$ using Adam optimizer to minimize $L_D$.
    \State $t \gets t + 1$ 

\EndWhile
\State \textbf{return} $\Theta_G, \Theta_D$
\end{algorithmic}
\end{algorithm}

In the generator training, the minimized total loss \(L_{G}\) is the summation of
the losses computed in \eqref{eq:adv_gen_loss}, \eqref{eq:time_loss}, \eqref{eq:freq_loss}, expressed as:
\begin{multline}
 L_{G} = (D(\tilde{x}) - 1)^2 + \lambda_{\text{time}} \cdot \| \tilde{x} - x \|_1 \\
 + \lambda_{\text{freq}} \cdot \left\| S(\tilde{x}) - S(x) \right\|_1
\label{eq:gen_loss}
\end{multline}

\subsubsection{Training the Discriminator Networks}
The training of the discriminator networks in the BRSR-OpGAN model adheres to the standard protocol used in generative adversarial networks (GANs). Unlike traditional GANs, which employ cross-entropy, we utilize Mean Squared Error (MSE) to calculate the adversarial
loss, effectively addressing the vanishing gradient problem often encountered during the GAN training.  For each pair of signals, consisting of a clean signal \(x\) and its corresponding noisy version \(\hat{x}\), the adversarial losses are computed as follows:
\begin{equation}
L_{adv}(D, \hat{x}, x) = (D(x) - 1)^2 + (D(\tilde{x}))^2 
\label{eq:disc_loss}
\end{equation}

In the following step, the discriminator weights are calculated by minimizing \(L_D = L_{adv}(D, \hat{x}, x) \). In this formulation, \(D(x)\) is the discriminator's assessment of the authenticity of the clean signal \(x\), which should ideally approach a value of 1, indicating \textit{real}. Conversely, \(D(\tilde{x})\) represents the discriminator's evaluation of the denoised signal produced by the generator \(G\), with the aim of this output approaching 0, indicating \textit{fake}. This setup ensures that the discriminator is tuned to distinguish effectively between the clean and restored signals.

In the training process, the generator and discriminator are trained adversarially, as described in Algorithm~\ref{alg:training}. One training iteration updates the generator model to fake the discriminator by forcing it to output 1 when fed with the restored signals. This behavior results from the minimization of \eqref{eq:adv_gen_loss}. Using the updated generator model, the restored signals are then computed again. The discriminators are then updated to produce the output of 1 for the clean signals, i.e., minimizing \eqref{eq:disc_loss}.

\subsection{Experimental Setup}
In our experiments, we adopt a training approach involving a maximum of 1000 back-propagation (BP) epochs and a batch size of 64. The Adam optimizer is employed, with the initial learning rates for both the generator and discriminator set at \(5 \cdot 10^{-4}\). Furthermore, we set the loss weights \(\lambda_{\text{time}}\) and \(\lambda_{\text{freq}}\) in \eqref{eq:gen_loss} to 100 and 200, respectively. We ensured consistency in network architecture and training parameter settings across all radar restoration experiments on both datasets. The order of the Taylor polynomials in all Self-ONN layers, Q, is set as 3 for all BRSR-OpGAN configurations.

To implement the proposed BRSR-OpGAN architecture, we utilized the FastONN library \cite{malik2020fastonn}, built upon Python and PyTorch. This setup ensures consistency and reliability across our experiments, facilitating robust evaluation and comparison of results.

\begin{figure*}[t!]
    \centering
    \includegraphics[width=0.95\linewidth,keepaspectratio]{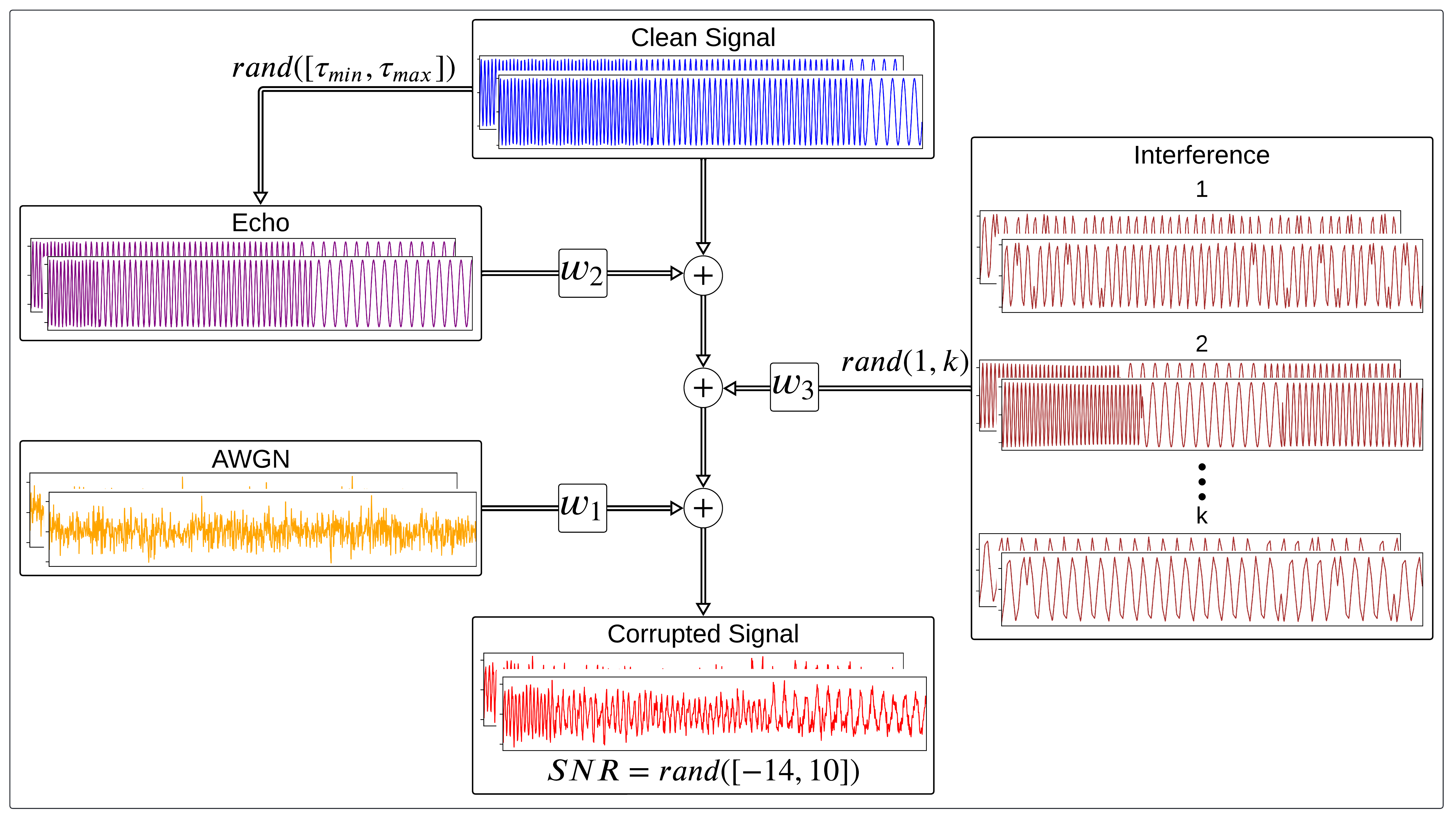}
    \caption{The illustration of extended real-world BRSR dataset generation with a random choice of artifacts and their severities.}
    \label{fig:figure_4}
\end{figure*}

\section{SIGNAL MODELING AND DATASET GENERATION}
\label{sec:dataset}
As discussed earlier, real-world radar signals are corrupted by various artifacts and noise when received, severely impacting their quality and interpretability. Understanding and modeling these distortions accurately is essential for developing and testing signal restoration algorithms under real-world conditions. This section presents the baseline dataset and the methodology for creating an extended radar dataset that mimics real-world situations by incorporating various artifacts with randomized parameters.

\subsection{Major Artifact Types of the Received Radar Signals}
The received radar signals are susceptible to various artifacts that can significantly degrade their quality and interpretability. This section outlines the major artifacts encountered in radar signal processing:

\subsubsection{Additive White Gaussian Noise (AWGN)}
Additive White Gaussian Noise (AWGN) is characterized by a constant power spectral density (PSD) across all frequencies. This type of noise can be mathematically modeled as:
\begin{equation}
\hat{x}(t) = x(t) + n(t)
\label{eq:noise_awgn}
\end{equation}
where \(\hat{x}(t)\) represents the received signal, \(x(t)\) is the original radar signal, and \(n(t)\) is the AWGN. The noise \(n(t)\) follows a Gaussian distribution with zero mean and variance \(\sigma^2\):

\subsubsection{Echo Signal Distortion}
Echo signal distortion occurs when a signal overlaps with its delayed replica:
\begin{equation}
\hat{x}(t) = x(t) + \alpha x(t-\tau)
\label{eq:noise_echo}
\end{equation}
where \(\alpha\) is the attenuation factor, and \(\tau\) is the delay of the echo. 

\subsubsection{Interference}
Interference arises when another signal interferes or disrupts the original radar signal:
\begin{equation}
\hat{x}(t) = x(t) + i(t) 
\label{eq:noise_interefernce}
\end{equation}
where \(i(t)\) is the interfering signal.

\subsubsection{Composite Signal Distortion}
This complex model combines all major artifacts with randomly assigned weights based on the desired SNR value as follows:
\begin{equation}
\hat{x}(t) = x(t) + w_1 \cdot n(t) + w_2 \cdot x(t-\tau) + w_3 \cdot i(t) 
\end{equation}

\subsection{Baseline Dataset}
As part of our study, we utilize the 12-class Signal Dataset proposed by the authors in \cite{jiang2024multilayer} as the baseline dataset. This dataset covers a range of SNR from -14 dB to 10 dB, with steps of 2 dB. Each signal is corrupted by adding AWGN as the only artifact, resulting in the generation of 400 samples at each SNR level with a total number of 62,400 samples. The dataset is divided into an 8:2 ratio, with 80\% of the samples assigned for training and 20\% for validation. To ensure the reliability of the training results, 150 test samples are generated for each signal at each SNR, resulting in a total of 23,400 signals for testing.

\subsection{BRSR Benchmark Dataset Generation}
In this study, we aim to generate radar modulation samples that simulate the complex and varied conditions encountered in real-world radar environments. Fig.~\ref{fig:figure_4} illustrates the methodology used to create the BRSR dataset that can accurately mimic actual received radar signals corrupted by any blend of common artifacts. To accomplish this, we corrupt clean (transmitted) radar signals with three major artifact types: Additive White Gaussian Noise (AWGN), Unwanted Echo, and Interference.

Each artifact type (AWGN, Echo, Interference) is additive, allowing the use of random weights to control its impact. This enables the creation of diverse scenarios by varying the severity of each artifact. Additionally, scenarios with either single or dual artifacts active are included to represent conditions where one or more types of distortions are absent so that a wide range of real-world signal corruptions can be simulated. 

We utilize 85,800 clean data samples from the baseline dataset to generate the BRSR dataset, which is also split into training, validation, and test sets, maintaining the same proportions as the baseline dataset. Each sample consists of a 2x1024 segment, with two channels representing the real and imaginary parts of the complex-valued radar signals. All signals are processed through the real-world corrupted radar generation setup (illustrated in Fig.~\ref{fig:figure_4}). The final dataset includes 85,800 samples with a mix of all artifact types, referred to as the BRSR dataset throughout this paper. Henceforth, this benchmark dataset can serve as a robust platform for developing and evaluating radar signal restoration techniques.

\subsection{Mathematical Modeling}
The generation of the BRSR dataset that simulates real-world radar signal conditions involves introducing the described artifacts into clean radar signals with randomized parameters. The algorithm for dataset creation is presented in Algorithm~\ref{alg:dataset_generation}. Given a clean radar signal \( x(t) \) from the baseline dataset, the process of the signal corruption with a blend of artifacts is modeled as follows:

\begin{enumerate}
    \item \textbf{Random SNR Assignment:} For each selected signal \( x(t) \), the desired SNR level is randomly chosen within the range of $-14$ dB to $10$ dB, represented as $\text{SNR}_{\text{rand}}$.

    \item \textbf{Calculation of Desired Noise Power:} Desired noise power is calculated as
    \begin{equation}
    P_{\text{rand}} = \frac{P_s}{10^{\text{SNR}_{\text{rand}} / 10}}
    \label{eq:noise_power}
    \end{equation}
    where \( P_s \) is the power of the clean signal \( x(t) \).
    \item \textbf{Selection of Artifacts:} The dataset generation algorithm randomly selects a combination of artifacts from the set \{$\text{AWGN}$, $\text{Echo}$, $\text{Intereference}$\}, resulting in seven possible scenarios, including individual artifacts, pairs, and the trio.

    \item \textbf{Echo Delay:} For echo artifacts, a delay parameter $\tau$ is randomly set within the predefined range, [128, 512].

    \item \textbf{Interference Signal Selection:} A signal from a predefined set of 100 signals is randomly chosen to simulate interference.

    \item \textbf{Artifact Weighting:} When multiple artifacts are selected, random pseudo weights $\tilde{w}_i$ are assigned to each artifact such that
    \begin{equation}
    \sum_{i=1}^{n} \tilde{w}_i = 1
    \label{eq:pseudo_weights}
    \end{equation}
    where $\tilde{w}_i$ is set to zero for the artifacts not present in the selected set.

    \item \textbf{Calculation of Actual Weights:} Actual weights are calculated using
    \begin{equation}
        w_i = \tilde{w}_i \times \sqrt{\frac{P_{\text{rand}}}{P_i}}
    \label{eq:actual_weights}
    \end{equation}
    where \( P_i \) is the actual noise power of the corresponding artifact. This equation ensures a proportional blend of artifacts based on the overall $\text{SNR}_{\text{rand}}$.

    \item \textbf{Distorted Signal:} The final degraded signal $\hat{x}(t)$ is obtained by applying the selected artifacts to the clean signal \( x(t) \) according to their respective weights and parameters. This can be mathematically expressed as:
    \begin{equation}
    \hat{x}(t) = x(t) + \sum_{i=1}^{n} w_i \cdot a_i(t)
    \label{eq:distorted_signal}
    \end{equation}
    where \( a_i(t) \) represents the individual artifacts.
\end{enumerate}

\begin{algorithm}[t!]
\caption{BRSR Dataset Generation with Randomized Artifacts}
\label{alg:dataset_generation}
\small
\begin{algorithmic}[1]
\State \textbf{Input:} Set of clean radar signals \(\{x(t)\}\), desired SNR range \([-14, 10]\) dB, echo delay range \([\tau_{\min}, \tau_{\max}]\), set of interference signals \(\{i_k(t)\}\)
\State \textbf{Output:} Pairs of clean and distorted signals \(\{(x(t), \hat{x}(t))\}\)
\For{each clean radar signal \(x(t)\)}
    \State Randomly select a desired SNR value \(\text{SNR}_{\text{rand}} \in [-14, 10]\) dB;
    \State Compute desired noise power \(P_n\) using \eqref{eq:noise_power};
    \State Randomly choose a combination of artifacts \(A \subset \{\text{AWGN}, \text{Echo}, \text{Interference}\}\);
    \For{each artifact \(a_i \in A\)}
        \State Generate artifact using \eqref{eq:noise_awgn}, \eqref{eq:noise_echo}, \eqref{eq:noise_interefernce};
    \EndFor
    \State Assign random pseudo weights using \eqref{eq:pseudo_weights};
    \State Calculate actual weights using \eqref{eq:actual_weights};
    
    \State Combine artifacts to form the distorted signal using \eqref{eq:distorted_signal}:
    \[
    \hat{x}(t) = x(t) + \sum_{i=1}^{n} w_i \cdot a_i(t)
    \]
    \State Save the pair \((x(t), \hat{x}(t))\)
\EndFor
\end{algorithmic}
\end{algorithm}

This process ensures the generation of a comprehensive dataset that accurately represents the variety and unpredictability of real-world radar signal conditions. As a result, each signal in this dataset is subjected to a unique combination of artifacts.

\section{EXPERIMENTAL RESULTS}
\label{sec:results}
In this section, we first present the performance evaluation metrics. Then, we evaluate the proposed BRSR-OpGAN model's performance over the baseline and extended BRSR datasets. Subsequently, a comparative analysis is presented, followed by a thorough discussion of the results obtained. We conclude this section with both quantitative and qualitative assessments and a detailed examination of the computational efficiency of the proposed approach.

\subsection{Evaluation Metrics}
The performance of signal restoration can be evaluated using several metrics, including SNR, Peak Signal-to-Noise Ratio (PSNR), and Mean Squared Error (MSE). These metrics provide a quantitative assessment of the restoration quality. The definitions and mathematical formulations of these metrics are as follows:

\subsubsection{SNR}
The SNR is used to measure the performance of signal restoration. It is defined as:
\begin{equation}
\text{SNR}_{\text{out}} = 10 \log_{10} \left( \frac{\sum_{i=1}^N x(t_i)^2}{\sum_{i=1}^N [x(t_i) - \tilde{x}(t_i)]^2} \right)
\label{eq:snr_out}
\end{equation}
where \( x(t) \) and \(\tilde{x}(t)\) represent the original clean signal and the restored signal, respectively, and \( N \) is the number of sampling points.

\subsubsection{PSNR}
The PSNR is another metric for assessing signal restoration quality, particularly in peak error. It is defined as:
\begin{equation}
\text{PSNR}_{\text{out}} = 10 \log_{10} \left( \frac{P_{\text{max}}^2}{\frac{1}{N}\sum_{i=1}^N [x(t_i) - \tilde{x}(t_i)]^2} \right)
\label{eq:psnr_out}
\end{equation}
where \( P_{\text{max}} \) is the maximum possible signal value in the original signal \( x(t) \), and \( \frac{1}{N}\sum_{i=1}^N [x(t_i) - \tilde{x}(t_i)]^2 \) represents the mean squared error (MSE) between the clean and the restored signals.

\subsubsection{MSE}
The MSE measures the average squared difference between the original (transmitted) and restored signals. It is expressed by:
\begin{equation}
\text{MSE} = \frac{1}{N}\sum_{i=1}^N [x(t_i) - \tilde{x}(t_i)]^2
\label{eq:mse_out}
\end{equation}

The standard metrics, SNR, PSNR, and MSE, collectively provide a comprehensive evaluation of the signal restoration performance, allowing us to quantify the performance of the restoration algorithms under various conditions.

\begin{table}[t!]
\centering
\caption{SNR values by different restoration methods over the baseline dataset.}
\label{tab:snr_comparison}
\resizebox{\linewidth}{!}{
\begin{tabular}{@{}lccccc@{}}
\toprule
\textbf{Restoration Algorithms} & \multicolumn{5}{c}{\textbf{Average SNR (dB)}} \\ \midrule
\cellcolor[gray]{.93}Corrupted Signal (Reference) & \cellcolor[gray]{.93}-14   & \cellcolor[gray]{.93}-8    & \cellcolor[gray]{.93}-2    & \cellcolor[gray]{.93}4     & \cellcolor[gray]{.93}10    \\
EEMD                 & -11.57 & -5.53 & 0.21  & 4.23  & 10.79 \\
\cellcolor[gray]{.93}VMD-WT               & \cellcolor[gray]{.93}-2.42  & \cellcolor[gray]{.93}-0.24 & \cellcolor[gray]{.93}3.75  & \cellcolor[gray]{.93}8.29  & \cellcolor[gray]{.93}13.03 \\
VMD-LMD-WT           & -2.24  & 0.48  & 4.77  & 9.51  & 14.31 \\
\cellcolor[gray]{.93}CGAN                 & \cellcolor[gray]{.93}1.29   & \cellcolor[gray]{.93}6.00  & \cellcolor[gray]{.93}11.40 &\cellcolor[gray]{.93}15.71  & \cellcolor[gray]{.93}18.64 \\
BRSR-OpGAN           & \textbf{1.31}   & \textbf{6.72}  & \textbf{12.11} & \textbf{16.53}  & \textbf{19.49} \\ \bottomrule
\end{tabular}
}
\label{tab:results_baseline}
\end{table}

\subsection{Quantitative Evaluations}
The quantitative evaluations aim to assess the effectiveness of the BRSR-OpGAN model in enhancing radar signal quality. We first utilize a baseline dataset involving radar signals corrupted only with AWGN to compare our model with the recent denoising methods. This setup allows us to benchmark our framework against traditional methods using published datasets under controlled conditions, highlighting the potential improvements offered by our approach. As the main objective of this study, we then introduce results over the extended BRSR dataset comprising more challenging real-world radar data with various artifacts to test the true performance of the proposed blind restoration approach.

For comparative evaluations and benchmarking, over the baseline dataset we used three recent denoising algorithms: ensemble empirical mode decomposition (EEMD) \cite{wu2009ensemble}, the combination of variational mode
decomposition and wavelet denoising algorithm (VMD-WT) \cite{qi2022method}, and a multilayer denoising method termed as VMD-LMD-WT \cite{jiang2024multilayer}. Despite the existence of numerous GAN-based methods for radar signal enhancement, most focus primarily on denoising and often assume a limited number of degradation sources with predefined SNR levels. A significant issue in the current literature is the lack of publicly available datasets, as most existing approaches do not share their data. Furthermore, these studies use a single noise type such as AWGN and evaluate their results based on the classification accuracy before and after restoration rather than the restoration performance. Consequently, it was challenging to utilize pre-trained weights, particularly for the BRSR dataset with an unknown blend of several artifacts. To address this limitation and facilitate comparative evaluations, we also conducted experiments using the conventional (CNN-based) GAN (CNN-GAN) with the same architecture as BRSR-OpGAN. This allows us to incorporate deep learning-based methods into our results, comprehensively evaluating radar signal restoration techniques.
 
In TABLE \ref{tab:results_baseline}, we present the performance of the proposed BRSR-OpGAN against the three recent denoising methods on the baseline dataset. Average SNR values are computed over all independent test samples at 5 different SNR levels. As the input SNR increases, the improvement in SNR generally decreases. This is typical in denoising applications because higher initial SNR values leave less room for enhancement. Notably, at an SNR of -14 dB, the SNR improvement peaks. At this level, the proposed BRSR-OpGAN method enhances the SNR by 15.31 dB, significantly surpassing the performance of the other recent denoising methods. It is interesting that, three competing methods show limited performance improvements despite the fact that AWGN noise is the only artifact corrupting the radar signals at predefined SNR levels, which aligns with our earlier discussion. This performance confirms our approach's efficacy in managing a comparatively simpler noise scenario and sets a solid baseline for further evaluations.

\begin{table*}[t!]
\centering
\caption{Performance comparison of different restoration methods over the BRSR dataset.}
\label{tab:results}
\resizebox{\linewidth}{!}{
\begin{tabular}{lccc}
    \toprule
    \textbf{Restoration Algorithms} & \textbf{Average SNR (dB)} & \textbf{Average PSNR (dB)} & \textbf{Average MSE} \\
    \midrule
    \cellcolor[gray]{.93}Corrupted Signal (Reference) & \cellcolor[gray]{.93}-1.94 & \cellcolor[gray]{.93}-1.27 & \cellcolor[gray]{.93}9.65 \\

    VMD-LMD-WT \cite{jiang2024multilayer} & 3.36 & 4.04 & 3.25 \\

    \cellcolor[gray]{.93}CNN-GAN (Time Domain) & \cellcolor[gray]{.93}8.86 & \cellcolor[gray]{.93}9.54 & \cellcolor[gray]{.93}0.39 \\

    CNN-GAN (Dual Domain) & 9.04 & 9.71 & 0.42 \\

    \cellcolor[gray]{.93}BRSR-OpGAN (Q=3, Time Domain) & \cellcolor[gray]{.93}9.53 & \cellcolor[gray]{.93}10.20 & \cellcolor[gray]{.93}0.34 \\

    BRSR-OpGAN (Q=3, Dual Domain) & 10.30 & 10.98 & 0.30 \\

    \cellcolor[gray]{.93}BRSR-OpGAN (Q=3, Dual Domain, 2nd Pass) & \cellcolor[gray]{.93}\textbf{12.36} & \cellcolor[gray]{.93}\textbf{13.03} & \cellcolor[gray]{.93}\textbf{0.22} \\
    \bottomrule
\end{tabular}
}
\label{tab:results_extended}
\end{table*}

Over the BRSR dataset, Table~\ref{tab:results_extended} summarizes the average SNR, PSNR, and MSE values for both corrupted and restored signals under different restoration algorithms. As discussed earlier, these are realistic restoration results since the restoration methods are blindly applied to the radar signals corrupted by unknown artifacts with randomized types and severities. 

Starting with multiple decomposition-based denoising (VMD-LMD-WT), we observe minimal improvement compared to the noisy reference signal. However, the CNN-GAN only with the temporal loss function shows significant improvement in the SNR, PSNR, and MSE metrics. This model serves as a solid baseline, achieving an average SNR of 8.86 dB and an average PSNR of 9.54 dB.

Further enhancements are established when using the CNN-GAN model with dual domain loss (both temporal and spectral losses). The average SNR increases to 9.04 dB, and the PSNR reaches 9.71 dB, with a slightly increased MSE of 0.42. This highlights the advantage of leveraging time and frequency domain information during training.

The proposed BRSR-OpGAN using only the time domain loss further improves performance compared to the CNN-GAN. The average SNR and PSNR values increase to 9.53 dB and 10.20 dB, respectively, and the MSE further decreases to 0.34. However, when the BRSR-OpGAN is trained with dual domain loss, it achieves state-of-the-art (SOTA) performance levels, with an average SNR of 10.30 dB, PSNR of 10.98 dB, and an MSE of 0.30.

The most significant improvements are observed with the BRSR-OpGAN (Q=3, Dual Domain, 2nd Pass) configuration. By utilizing a second-pass BRSR-OpGAN training, where the restored signal from the first pass is used as the input, the second restoration pass by the generator of the second BRSR-OpGAN achieves the highest average SNR of 12.36 dB and PSNR of 13.03 dB, with the lowest MSE of 0.22. Such a two-pass restoration approach significantly boosts the model's performance and achieves the highest restoration performance of above 14 dB average SNR improvement.

\begin{figure}[!t]
    \centering
    \includegraphics[width=0.94\linewidth]{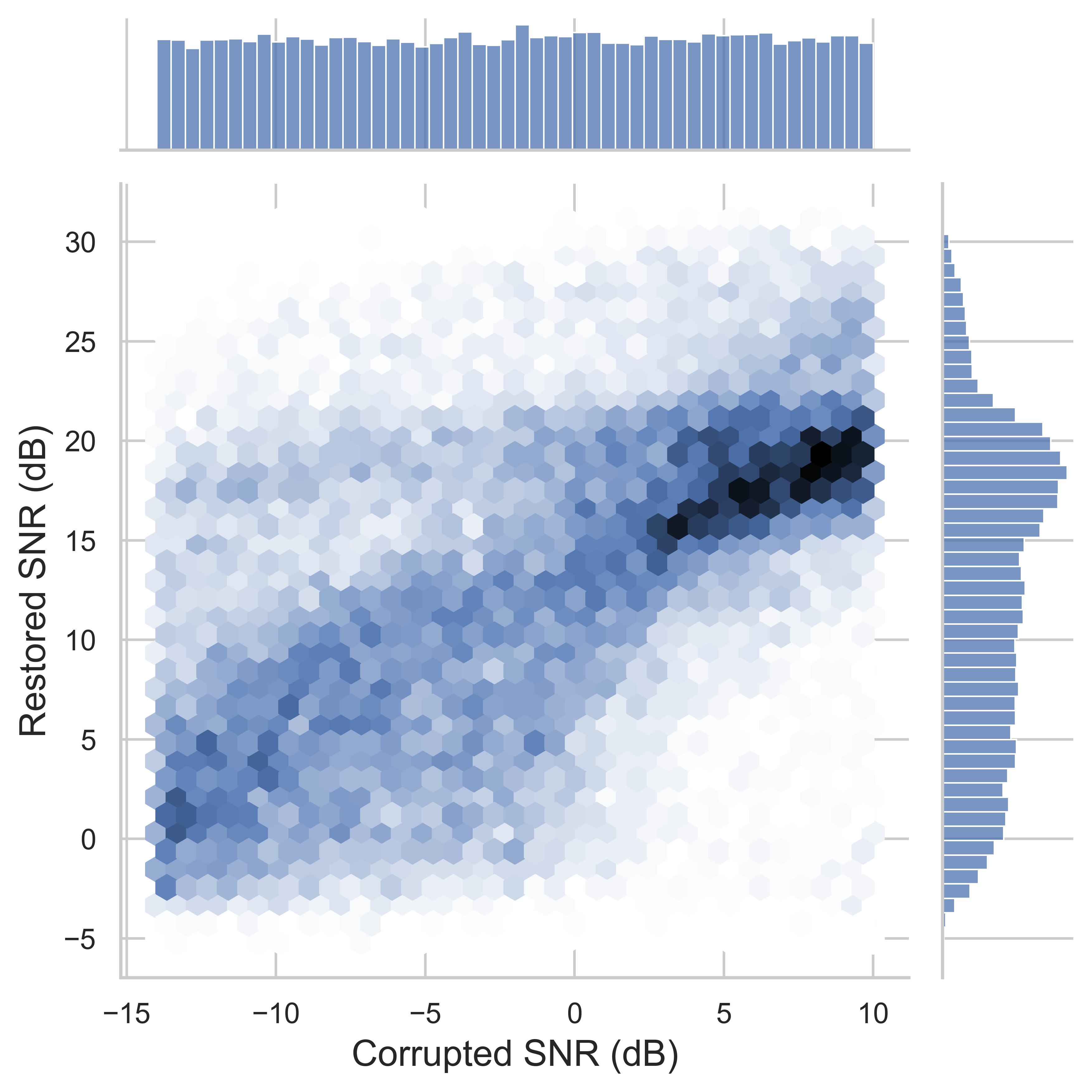}
    \caption{Joint distribution of corrupted vs restored SNR.}
    \label{fig:SNR_joint_distribution}
\end{figure}

Fig.~\ref{fig:SNR_joint_distribution} illustrates the SNR of all test data signals before and after restoration using the BRSR-OpGAN model. The joint distribution highlights the relationship between corrupted SNR values (x-axis) and restored SNR values (y-axis). Notably, even for the corrupted signals with the lowest SNR values, down to -14 dB with a randomized blend of artifacts, the model shows a significant improvement in all SNR levels. Additionally, the marginal histograms on the top and right sides of the plot display the distribution of corrupted and restored SNR values, respectively. The restored SNR values are skewed towards higher ranges, confirming the model’s robustness in enhancing signal quality under challenging conditions.

\begin{figure}[t]
    \centering
    \includegraphics[width=0.94\linewidth]{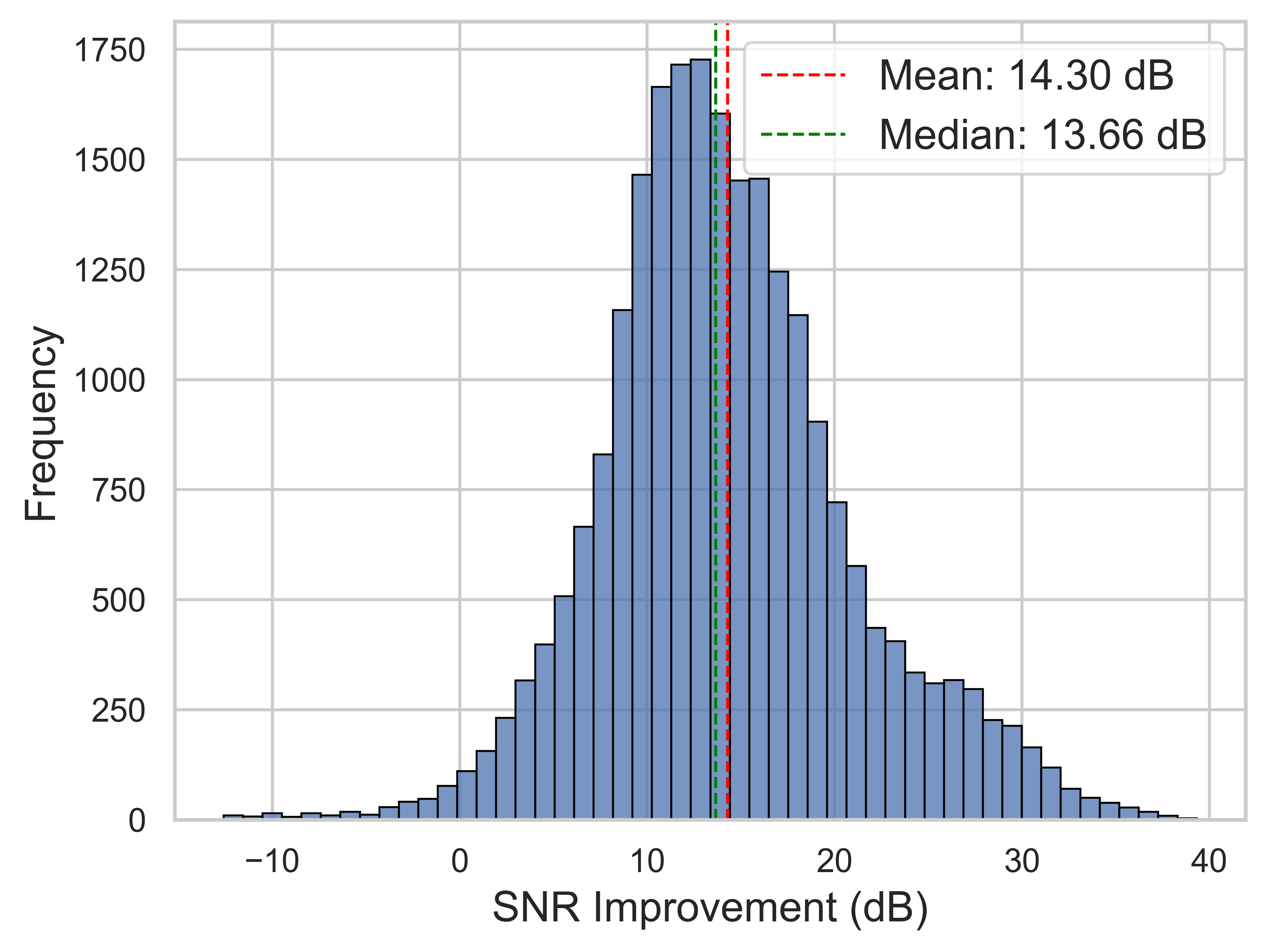}
    \caption{Histogram of SNR improvements.}
    \label{fig:SRN_improvement_histogram}
\end{figure}

Fig.~\ref{fig:SRN_improvement_histogram} presents the histogram of the SNR improvement achieved by the BRSR-OpGAN model across the test dataset. The SNR improvement values follow a roughly normal distribution centered around a mean improvement of 14.30 dB, as indicated by the red dashed line, with the median slightly lower at 13.66 dB, marked by the green dashed line. This close alignment suggests a symmetrical distribution of SNR improvement values. The peak of the histogram occurs around the mean, showing that most test signals experienced an SNR improvement of approximately 14 dB. The spread of the data, with improvements ranging from slightly negative values to over 30 dB, demonstrates the model's robustness in handling a wide variety of corrupted signals. While a few instances show negative SNR improvement, indicating some severe artifacts corrupting the signal beyond any restoration; however, their occurrence is minimal. The substantial mean and median SNR improvements highlight the model's effectiveness in enhancing signal quality, confirming its potential for reliable blind radar signal restoration in practical applications. Overall, the significant SNR enhancements validate the BRSR-OpGAN model's capability to restore radar signals effectively in real-world scenarios.

These findings underscore the importance of dual-domain loss function during training and iterative enhancement techniques in improving radar signal restoration. The 2-pass restoration approach, particularly, shows substantial promise in achieving superior restoration quality, setting a new benchmark for future research in this domain. The BRSR-OpGAN (Q=3, Dual Domain, 2nd Pass) model is the most effective configuration, providing a robust high-fidelity radar signal restoration framework.

\subsection{Qualitative Evaluation}
\begin{figure*}[!htb]
\centering
\includegraphics[width=\textwidth,keepaspectratio]{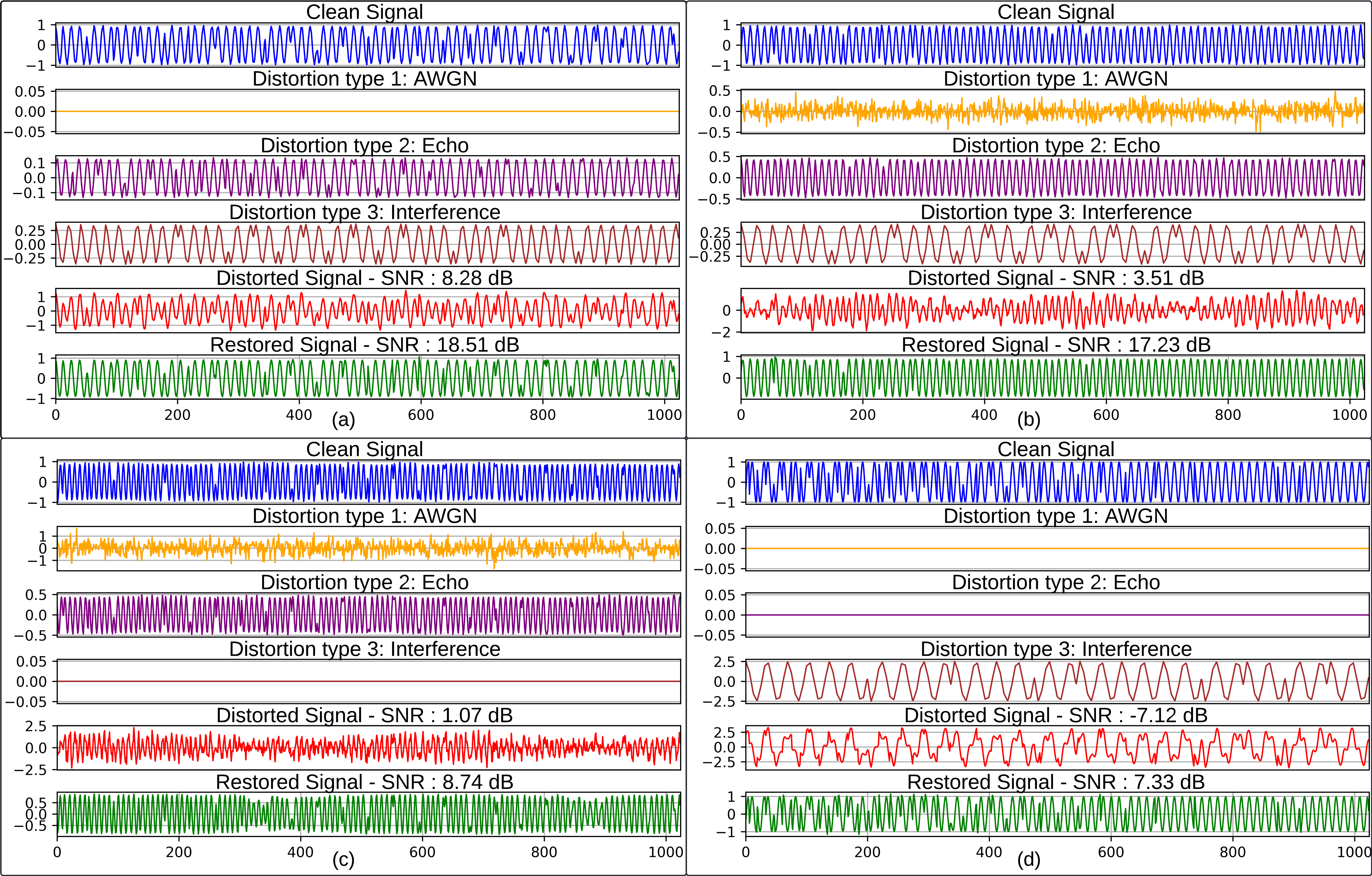}
\caption{Sample radar signals corrupted with random blend artifacts and the corresponding BRSR-OpGAN restored signal.}
\label{fig:qualitative}
\end{figure*}

Fig. \ref{fig:qualitative} illustrates examples of radar signal restoration by BRSR-OpGAN. In each figure, the top plot displays the original (transmitted) radar signal \(x(t)\) followed by a sequence of three plots, each exhibiting individual components of the major artifacts, AWGN, Echo, and Interference. The fifth plot shows the overall distorted signal \(\tilde{x}(t)\), which combines the artifacts, and the bottom plot demonstrates the restored signal \(\hat{x}(t)\) produced by the BRSR-OpGAN model.

\subsection{Computational Complexity}
In order to determine the computational complexity of restoring a radar signal (2x1024 samples) in each network configuration, we calculate the number of parameters (PARs) and the amount of inference time required for a batch size of 64. The detailed formulations for PARs calculations in Self-Organizing Neural Networks (Self-ONNs) are provided in \cite{malik2020fastonn}. All experiments were executed on a system equipped with a 2.4 GHz Intel Core i7 processor, 32 GB of RAM, and NVIDIA GeForce RTX A1000 6 GB graphics card. The BRSR-OpGAN was implemented using Python with the PyTorch library, and the classifier's training and testing phases were performed on GPU cores. Notably, the inference time was also measured using only a single CPU. As detailed in TABLE 4, the BRSR-OpGAN network needs more multiply-accumulate operations. However, a large proportion of these operations can be performed independently and in parallel. Consequently, an efficient implementation utilizing the vectorized formulation \cite{gabbouj2022robust} results in an actual running time that is only marginally higher than the equivalent CNN-based GAN. This increase is negligible when considering the significant gain in the restoration performance.

\begin{table}
\centering
\caption{Models and Their Computational Complexities}
\label{tab:performance_comparison}
\resizebox{\columnwidth}{!}{
\begin{tabular}{lcccccc}
    \toprule
    \textbf{Model} & \multicolumn{3}{c}{\textbf{Parameters (PARs) (K)}} & \multicolumn{2}{c}{\textbf{Inference Time (ms)}} \\
    \cmidrule(lr){2-4} \cmidrule(lr){5-6}
    & \textbf{G} & \textbf{D} & \textbf{Total} & \textbf{CPU} & \textbf{GPU} \\
    \midrule
    BRSR-OpGAN & 275.52 & 82.61 & 358.13 & 0.710 & 0.039 \\
    CNN-GAN    & 103.30 & 30.19 & 133.49 & 0.440 & 0.036 \\
    \bottomrule
\end{tabular}
}
\end{table}

\section{CONCLUSION}
\label{sec:conclusion}
The main challenge in real-world radar signal restoration is the degradation of acquired radar signals by various artifacts such as echo, sensor noise, signal jamming, and atmospheric disturbances. These artifacts can vary greatly in type, intensity, and duration, leading to dynamically and non-uniformly corrupted signals that are difficult to clean effectively. Despite extensive research in radar signal denoising and interference mitigation, there has not been a comprehensive solution targeting these multifaceted challenges in blind radar signal restoration.

In this context, this study introduces a pioneering approach that employs 1D Operational GANs iteratively for the blind restoration of real-world radar signals. The proposed model was evaluated using a benchmark radar signal dataset and an extended BRSR dataset by incorporating all major artifact types in a randomized manner, thus mimicking real-world scenarios. Our approach diverges from traditional methods by not making any prior assumptions about the type or severity of the artifacts corrupting the signal.

We developed the BRSR-OpGAN model as a compact, efficient end-to-end system for radar signal restoration, introducing a unique loss function for the generator that leverages both temporal and spectral characteristics of radar signals. This enables our model to adeptly suppress various artifacts while maintaining the integrity and characteristics of the original signal. Our model demonstrates unprecedented restoration performance over the most recent methods, indicating its potential for enhancing the clarity and usability of severely corrupted radar signals across various applications.

Finally, the proposed approach has an elegant computational efficiency, which further underscores its effectiveness, making it suitable for real-time applications on resource-constrained platforms. Future research will explore further optimizations and extensions of the proposed Op-GAN model, i) to incorporate the classification objective into the restoration to restore the signals to maximize the classification performance, and ii) to handle even more complex scenarios and improve robustness against a wider variety of signal distortions.

\section*{Acknowledgments}
This work was supported by the Business Finland project AMaLIA and NSF Center for Big Learning (CBL).

\section*{Data and Code Availability}
We have made the source code and the real-world radar dataset publicly available in a GitHub repository at \url{https://github.com/MUzairZahid/BRSR-OpGAN} to facilitate further research.


\bibliographystyle{cas-model2-names}

\bibliography{refs}

\begin{thebibliography}{32}
\expandafter\ifx\csname natexlab\endcsname\relax\def\natexlab#1{#1}\fi
\providecommand{\url}[1]{\texttt{#1}}
\providecommand{\href}[2]{#2}
\providecommand{\path}[1]{#1}
\providecommand{\DOIprefix}{doi:}
\providecommand{\ArXivprefix}{arXiv:}
\providecommand{\URLprefix}{URL: }
\providecommand{\Pubmedprefix}{pmid:}
\providecommand{\doi}[1]{\href{http://dx.doi.org/#1}{\path{#1}}}
\providecommand{\Pubmed}[1]{\href{pmid:#1}{\path{#1}}}
\providecommand{\bibinfo}[2]{#2}
\ifx\xfnm\relax \def\xfnm[#1]{\unskip,\space#1}\fi
\bibitem[{Dee()}]{DeepSig}
, .
\newblock \bibinfo{title}{Datasets - deepsig}.
\newblock \URLprefix \url{https://www.deepsig.ai/datasets/}.
\bibitem[{De~Martino(2018)}]{de2018introduction}
\bibinfo{author}{De~Martino, A.}, \bibinfo{year}{2018}.
\newblock \bibinfo{title}{Introduction to modern EW systems}.
\newblock \bibinfo{publisher}{Artech house}.
\bibitem[{Du et~al.(2022)Du, Zhong, Cai and Bi}]{du2022dncnet}
\bibinfo{author}{Du, M.}, \bibinfo{author}{Zhong, P.}, \bibinfo{author}{Cai, X.}, \bibinfo{author}{Bi, D.}, \bibinfo{year}{2022}.
\newblock \bibinfo{title}{Dncnet: Deep radar signal denoising and recognition}.
\newblock \bibinfo{journal}{IEEE Transactions on Aerospace and Electronic Systems} \bibinfo{volume}{58}, \bibinfo{pages}{3549--3562}.
\bibitem[{Fuchs et~al.(2020)Fuchs, Dubey, L{\"u}bke, Weigel and Lurz}]{fuchs2020automotive}
\bibinfo{author}{Fuchs, J.}, \bibinfo{author}{Dubey, A.}, \bibinfo{author}{L{\"u}bke, M.}, \bibinfo{author}{Weigel, R.}, \bibinfo{author}{Lurz, F.}, \bibinfo{year}{2020}.
\newblock \bibinfo{title}{Automotive radar interference mitigation using a convolutional autoencoder}, in: \bibinfo{booktitle}{2020 IEEE International Radar Conference (RADAR)}, \bibinfo{organization}{IEEE}. pp. \bibinfo{pages}{315--320}.
\bibitem[{Gabbouj et~al.(2022)Gabbouj, Kiranyaz, Malik, Zahid, Ince, Chowdhury, Khandakar and Tahir}]{gabbouj2022robust}
\bibinfo{author}{Gabbouj, M.}, \bibinfo{author}{Kiranyaz, S.}, \bibinfo{author}{Malik, J.}, \bibinfo{author}{Zahid, M.U.}, \bibinfo{author}{Ince, T.}, \bibinfo{author}{Chowdhury, M.E.}, \bibinfo{author}{Khandakar, A.}, \bibinfo{author}{Tahir, A.}, \bibinfo{year}{2022}.
\newblock \bibinfo{title}{Robust peak detection for holter ecgs by self-organized operational neural networks}.
\newblock \bibinfo{journal}{IEEE Transactions on Neural Networks and Learning Systems} \bibinfo{volume}{34}, \bibinfo{pages}{9363--9374}.
\bibitem[{Goodfellow et~al.(2014)Goodfellow, Pouget-Abadie, Mirza, Xu, Warde-Farley, Ozair, Courville and Bengio}]{goodfellow2014generative}
\bibinfo{author}{Goodfellow, I.}, \bibinfo{author}{Pouget-Abadie, J.}, \bibinfo{author}{Mirza, M.}, \bibinfo{author}{Xu, B.}, \bibinfo{author}{Warde-Farley, D.}, \bibinfo{author}{Ozair, S.}, \bibinfo{author}{Courville, A.}, \bibinfo{author}{Bengio, Y.}, \bibinfo{year}{2014}.
\newblock \bibinfo{title}{Generative adversarial nets}.
\newblock \bibinfo{journal}{Advances in neural information processing systems} \bibinfo{volume}{27}.
\bibitem[{Ince et~al.(2022)Ince, Kiranyaz, Devecioglu, Khan, Chowdhury and Gabbouj}]{ince2022blind}
\bibinfo{author}{Ince, T.}, \bibinfo{author}{Kiranyaz, S.}, \bibinfo{author}{Devecioglu, O.C.}, \bibinfo{author}{Khan, M.S.}, \bibinfo{author}{Chowdhury, M.}, \bibinfo{author}{Gabbouj, M.}, \bibinfo{year}{2022}.
\newblock \bibinfo{title}{Blind restoration of real-world audio by 1d operational gans}.
\newblock \bibinfo{journal}{arXiv preprint arXiv:2212.14618} .
\bibitem[{Isola et~al.(2017)Isola, Zhu, Zhou and Efros}]{isola2017image}
\bibinfo{author}{Isola, P.}, \bibinfo{author}{Zhu, J.Y.}, \bibinfo{author}{Zhou, T.}, \bibinfo{author}{Efros, A.A.}, \bibinfo{year}{2017}.
\newblock \bibinfo{title}{Image-to-image translation with conditional adversarial networks}, in: \bibinfo{booktitle}{Proceedings of the IEEE conference on computer vision and pattern recognition}, pp. \bibinfo{pages}{1125--1134}.
\bibitem[{Jan and Pietrow(2020)}]{jan2020artificial}
\bibinfo{author}{Jan, M.}, \bibinfo{author}{Pietrow, D.}, \bibinfo{year}{2020}.
\newblock \bibinfo{title}{Artificial neural networks in the filtration of radiolocation information}, in: \bibinfo{booktitle}{2020 IEEE 15th International Conference on Advanced Trends in Radioelectronics, Telecommunications and Computer Engineering (TCSET)}, \bibinfo{organization}{IEEE}. pp. \bibinfo{pages}{680--685}.
\bibitem[{Jiang et~al.(2024)Jiang, Zhou, Shen, Wang, Quan and Jin}]{jiang2024multilayer}
\bibinfo{author}{Jiang, M.}, \bibinfo{author}{Zhou, F.}, \bibinfo{author}{Shen, L.}, \bibinfo{author}{Wang, X.}, \bibinfo{author}{Quan, D.}, \bibinfo{author}{Jin, N.}, \bibinfo{year}{2024}.
\newblock \bibinfo{title}{Multilayer decomposition denoising empowered cnn for radar signal modulation recognition}.
\newblock \bibinfo{journal}{IEEE Access} .
\bibitem[{Kiranyaz et~al.(2022)Kiranyaz, Devecioglu, Ince, Malik, Chowdhury, Hamid, Mazhar, Khandakar, Tahir, Rahman et~al.}]{kiranyaz2022blind}
\bibinfo{author}{Kiranyaz, S.}, \bibinfo{author}{Devecioglu, O.C.}, \bibinfo{author}{Ince, T.}, \bibinfo{author}{Malik, J.}, \bibinfo{author}{Chowdhury, M.}, \bibinfo{author}{Hamid, T.}, \bibinfo{author}{Mazhar, R.}, \bibinfo{author}{Khandakar, A.}, \bibinfo{author}{Tahir, A.}, \bibinfo{author}{Rahman, T.}, et~al., \bibinfo{year}{2022}.
\newblock \bibinfo{title}{Blind ecg restoration by operational cycle-gans}.
\newblock \bibinfo{journal}{IEEE Transactions on Biomedical Engineering} \bibinfo{volume}{69}, \bibinfo{pages}{3572--3581}.
\bibitem[{Kiranyaz et~al.(2017a)Kiranyaz, Ince, Iosifidis and Gabbouj}]{kiranyaz2017generalized}
\bibinfo{author}{Kiranyaz, S.}, \bibinfo{author}{Ince, T.}, \bibinfo{author}{Iosifidis, A.}, \bibinfo{author}{Gabbouj, M.}, \bibinfo{year}{2017}a.
\newblock \bibinfo{title}{Generalized model of biological neural networks: Progressive operational perceptrons}, in: \bibinfo{booktitle}{2017 International Joint Conference on Neural Networks (IJCNN)}, \bibinfo{organization}{IEEE}. pp. \bibinfo{pages}{2477--2485}.
\bibitem[{Kiranyaz et~al.(2017b)Kiranyaz, Ince, Iosifidis and Gabbouj}]{kiranyaz2017progressive}
\bibinfo{author}{Kiranyaz, S.}, \bibinfo{author}{Ince, T.}, \bibinfo{author}{Iosifidis, A.}, \bibinfo{author}{Gabbouj, M.}, \bibinfo{year}{2017}b.
\newblock \bibinfo{title}{Progressive operational perceptrons}.
\newblock \bibinfo{journal}{Neurocomputing} \bibinfo{volume}{224}, \bibinfo{pages}{142--154}.
\bibitem[{Kiranyaz et~al.(2020)Kiranyaz, Ince, Iosifidis and Gabbouj}]{kiranyaz2020operational}
\bibinfo{author}{Kiranyaz, S.}, \bibinfo{author}{Ince, T.}, \bibinfo{author}{Iosifidis, A.}, \bibinfo{author}{Gabbouj, M.}, \bibinfo{year}{2020}.
\newblock \bibinfo{title}{Operational neural networks}.
\newblock \bibinfo{journal}{Neural Computing and Applications} \bibinfo{volume}{32}, \bibinfo{pages}{6645--6668}.
\bibitem[{Kiranyaz et~al.(2021a)Kiranyaz, Malik, Abdallah, Ince, Iosifidis and Gabbouj}]{kiranyaz2021exploiting}
\bibinfo{author}{Kiranyaz, S.}, \bibinfo{author}{Malik, J.}, \bibinfo{author}{Abdallah, H.B.}, \bibinfo{author}{Ince, T.}, \bibinfo{author}{Iosifidis, A.}, \bibinfo{author}{Gabbouj, M.}, \bibinfo{year}{2021}a.
\newblock \bibinfo{title}{Exploiting heterogeneity in operational neural networks by synaptic plasticity}.
\newblock \bibinfo{journal}{Neural Computing and Applications} \bibinfo{volume}{33}, \bibinfo{pages}{7997--8015}.
\bibitem[{Kiranyaz et~al.(2021b)Kiranyaz, Malik, Abdallah, Ince, Iosifidis and Gabbouj}]{kiranyaz2021self}
\bibinfo{author}{Kiranyaz, S.}, \bibinfo{author}{Malik, J.}, \bibinfo{author}{Abdallah, H.B.}, \bibinfo{author}{Ince, T.}, \bibinfo{author}{Iosifidis, A.}, \bibinfo{author}{Gabbouj, M.}, \bibinfo{year}{2021}b.
\newblock \bibinfo{title}{Self-organized operational neural networks with generative neurons}.
\newblock \bibinfo{journal}{Neural Networks} \bibinfo{volume}{140}, \bibinfo{pages}{294--308}.
\bibitem[{Kishore and Rao(2017)}]{kishore2017automatic}
\bibinfo{author}{Kishore, T.R.}, \bibinfo{author}{Rao, K.D.}, \bibinfo{year}{2017}.
\newblock \bibinfo{title}{Automatic intrapulse modulation classification of advanced lpi radar waveforms}.
\newblock \bibinfo{journal}{IEEE Transactions on Aerospace and Electronic Systems} \bibinfo{volume}{53}, \bibinfo{pages}{901--914}.
\bibitem[{Kong et~al.(2018)Kong, Kim, Hoang and Kim}]{kong2018automatic}
\bibinfo{author}{Kong, S.H.}, \bibinfo{author}{Kim, M.}, \bibinfo{author}{Hoang, L.M.}, \bibinfo{author}{Kim, E.}, \bibinfo{year}{2018}.
\newblock \bibinfo{title}{Automatic lpi radar waveform recognition using cnn}.
\newblock \bibinfo{journal}{Ieee Access} \bibinfo{volume}{6}, \bibinfo{pages}{4207--4219}.
\bibitem[{Lee et~al.(2019)Lee, Lee and Kim}]{lee2019mutual}
\bibinfo{author}{Lee, S.}, \bibinfo{author}{Lee, J.Y.}, \bibinfo{author}{Kim, S.C.}, \bibinfo{year}{2019}.
\newblock \bibinfo{title}{Mutual interference suppression using wavelet denoising in automotive fmcw radar systems}.
\newblock \bibinfo{journal}{IEEE Transactions on Intelligent Transportation Systems} \bibinfo{volume}{22}, \bibinfo{pages}{887--897}.
\bibitem[{Malik et~al.(2020)Malik, Kiranyaz and Gabbouj}]{malik2020fastonn}
\bibinfo{author}{Malik, J.}, \bibinfo{author}{Kiranyaz, S.}, \bibinfo{author}{Gabbouj, M.}, \bibinfo{year}{2020}.
\newblock \bibinfo{title}{Fastonn--python based open-source gpu implementation for operational neural networks}.
\newblock \bibinfo{journal}{arXiv preprint arXiv:2006.02267} .
\bibitem[{Mvuh et~al.(2024)Mvuh, Ebode~Ko’a and Bodo}]{mvuh2024multichannel}
\bibinfo{author}{Mvuh, F.L.}, \bibinfo{author}{Ebode~Ko’a, C.O.V.}, \bibinfo{author}{Bodo, B.}, \bibinfo{year}{2024}.
\newblock \bibinfo{title}{Multichannel high noise level ecg denoising based on adversarial deep learning}.
\newblock \bibinfo{journal}{Scientific Reports} \bibinfo{volume}{14}, \bibinfo{pages}{801}.
\bibitem[{Pace(2009)}]{pace2009detecting}
\bibinfo{author}{Pace, P.E.}, \bibinfo{year}{2009}.
\newblock \bibinfo{title}{Detecting and classifying low probability of intercept radar}.
\newblock \bibinfo{publisher}{Artech house}.
\bibitem[{Qi et~al.(2022)Qi, Wei, Feng, Zhang, Zhao and Guo}]{qi2022method}
\bibinfo{author}{Qi, T.}, \bibinfo{author}{Wei, X.}, \bibinfo{author}{Feng, G.}, \bibinfo{author}{Zhang, F.}, \bibinfo{author}{Zhao, D.}, \bibinfo{author}{Guo, J.}, \bibinfo{year}{2022}.
\newblock \bibinfo{title}{A method for reducing transient electromagnetic noise: Combination of variational mode decomposition and wavelet denoising algorithm}.
\newblock \bibinfo{journal}{Measurement} \bibinfo{volume}{198}, \bibinfo{pages}{111420}.
\bibitem[{Seddighi et~al.(2020)Seddighi, Ahmadzadeh and Taban}]{seddighi2020radar}
\bibinfo{author}{Seddighi, Z.}, \bibinfo{author}{Ahmadzadeh, M.R.}, \bibinfo{author}{Taban, M.R.}, \bibinfo{year}{2020}.
\newblock \bibinfo{title}{Radar signals classification using energy-time-frequency distribution features}.
\newblock \bibinfo{journal}{IET Radar, Sonar \& Navigation} \bibinfo{volume}{14}, \bibinfo{pages}{707--715}.
\bibitem[{Wang et~al.(2017)Wang, Wang and Zhang}]{wang2017automatic}
\bibinfo{author}{Wang, C.}, \bibinfo{author}{Wang, J.}, \bibinfo{author}{Zhang, X.}, \bibinfo{year}{2017}.
\newblock \bibinfo{title}{Automatic radar waveform recognition based on time-frequency analysis and convolutional neural network}, in: \bibinfo{booktitle}{2017 IEEE International Conference on Acoustics, Speech and Signal Processing (ICASSP)}, \bibinfo{organization}{IEEE}. pp. \bibinfo{pages}{2437--2441}.
\bibitem[{Wu and Huang(2009)}]{wu2009ensemble}
\bibinfo{author}{Wu, Z.}, \bibinfo{author}{Huang, N.E.}, \bibinfo{year}{2009}.
\newblock \bibinfo{title}{Ensemble empirical mode decomposition: a noise-assisted data analysis method}.
\newblock \bibinfo{journal}{Advances in adaptive data analysis} \bibinfo{volume}{1}, \bibinfo{pages}{1--41}.
\bibitem[{Zahid et~al.(2022)Zahid, Kiranyaz and Gabbouj}]{zahid2022global}
\bibinfo{author}{Zahid, M.U.}, \bibinfo{author}{Kiranyaz, S.}, \bibinfo{author}{Gabbouj, M.}, \bibinfo{year}{2022}.
\newblock \bibinfo{title}{Global ecg classification by self-operational neural networks with feature injection}.
\newblock \bibinfo{journal}{IEEE Transactions on Biomedical Engineering} \bibinfo{volume}{70}, \bibinfo{pages}{205--215}.
\bibitem[{Zahid et~al.(2021)Zahid, Kiranyaz, Ince, Devecioglu, Chowdhury, Khandakar, Tahir and Gabbouj}]{zahid2021robust}
\bibinfo{author}{Zahid, M.U.}, \bibinfo{author}{Kiranyaz, S.}, \bibinfo{author}{Ince, T.}, \bibinfo{author}{Devecioglu, O.C.}, \bibinfo{author}{Chowdhury, M.E.}, \bibinfo{author}{Khandakar, A.}, \bibinfo{author}{Tahir, A.}, \bibinfo{author}{Gabbouj, M.}, \bibinfo{year}{2021}.
\newblock \bibinfo{title}{Robust r-peak detection in low-quality holter ecgs using 1d convolutional neural network}.
\newblock \bibinfo{journal}{IEEE Transactions on Biomedical Engineering} \bibinfo{volume}{69}, \bibinfo{pages}{119--128}.
\bibitem[{Zhang et~al.(2017)Zhang, Diao and Guo}]{zhang2017convolutional}
\bibinfo{author}{Zhang, M.}, \bibinfo{author}{Diao, M.}, \bibinfo{author}{Guo, L.}, \bibinfo{year}{2017}.
\newblock \bibinfo{title}{Convolutional neural networks for automatic cognitive radio waveform recognition}.
\newblock \bibinfo{journal}{IEEE access} \bibinfo{volume}{5}, \bibinfo{pages}{11074--11082}.
\bibitem[{Zhang et~al.(2019)Zhang, Wang, Gan, Sun and Wang}]{zhang2019automatic}
\bibinfo{author}{Zhang, Z.}, \bibinfo{author}{Wang, C.}, \bibinfo{author}{Gan, C.}, \bibinfo{author}{Sun, S.}, \bibinfo{author}{Wang, M.}, \bibinfo{year}{2019}.
\newblock \bibinfo{title}{Automatic modulation classification using convolutional neural network with features fusion of spwvd and bjd}.
\newblock \bibinfo{journal}{IEEE Transactions on Signal and Information Processing over Networks} \bibinfo{volume}{5}, \bibinfo{pages}{469--478}.
\bibitem[{Zhou et~al.(2018)Zhou, Huang, Chen and Gao}]{zhou2018automatic}
\bibinfo{author}{Zhou, Z.}, \bibinfo{author}{Huang, G.}, \bibinfo{author}{Chen, H.}, \bibinfo{author}{Gao, J.}, \bibinfo{year}{2018}.
\newblock \bibinfo{title}{Automatic radar waveform recognition based on deep convolutional denoising auto-encoders}.
\newblock \bibinfo{journal}{Circuits, Systems, and Signal Processing} \bibinfo{volume}{37}, \bibinfo{pages}{4034--4048}.
\bibitem[{Zhou et~al.(2019)Zhou, Huang and Wang}]{zhou2019ensemble}
\bibinfo{author}{Zhou, Z.}, \bibinfo{author}{Huang, G.}, \bibinfo{author}{Wang, X.}, \bibinfo{year}{2019}.
\newblock \bibinfo{title}{Ensemble convolutional neural networks for automatic fusion recognition of multi-platform radar emitters}.
\newblock \bibinfo{journal}{ETRI Journal} \bibinfo{volume}{41}, \bibinfo{pages}{750--759}.

\end{thebibliography}





\end{document}